\documentclass[journal]{IEEEtran}

\ifCLASSINFOpdf
\else
\usepackage[dvips]{graphicx}
\fi

\usepackage{amssymb}

\usepackage[cmex10]{amsmath}

\DeclareMathOperator*{\argmin}{arg\,min}
\usepackage{tabularx}
\usepackage{booktabs}
\usepackage{multirow}
\usepackage{slashbox}
\usepackage{amsthm} 
\newtheorem{MyTheorem}{Theorem}

\usepackage{extarrows}
\usepackage{hyperref}
\usepackage{enumerate}
\usepackage[dvips]{graphicx}
\usepackage{subfig}
\usepackage{bm}
\usepackage{cite}
\usepackage{color}
\begin{document}

\title{Discriminative Block-Diagonal Representation Learning for Image Recognition}

\author{Zheng~Zhang, Yong~Xu,~\IEEEmembership{Senior Member,~IEEE}, Ling~Shao,~\IEEEmembership{Senior Member,~IEEE}, Jian~Yang,~\IEEEmembership{Member,~IEEE}
\thanks{Z. Zhang and Y. Xu are with the Bio-Computing Research Center, Shenzhen Graduate School, Harbin Institute of Technology, Shenzhen 518055, P. R. China (e-mail:darrenzz219@gmail.com; yongxu@ymail.com).}
\thanks{L. Shao is with the School of Computing Sciences, University of East Anglia, Norwich NR4 7TJ, U.K. (e-mail:E-mail: ling.shao@ieee.org.)}
\thanks{Jian Yang is with the College of Computer Science and Technology, Nanjing University of Science and Technology, Nanjing 210094, P. R. China (e-mail: csjyang@njust.edu.cn).}}

\markboth{IEEE Trans. NNLS,~June~2017}%
{Shell \MakeLowercase{\textit{et al.}}: Bare Demo of IEEEtran.cls for Journals}

\maketitle

\begin{abstract}
Existing block-diagonal representation researches mainly focuses on casting block-diagonal regularization on training data, while only little attention is dedicated to concurrently learning both block-diagonal representations of training and test data. In this paper, we propose a discriminative block-diagonal low-rank representation (BDLRR) method for recognition. In particular, the elaborate BDLRR is formulated as a joint optimization problem of shrinking the unfavorable representation from off-block-diagonal elements and strengthening the compact block-diagonal representation under the semi-supervised framework of low-rank representation. To this end, we first impose penalty constraints on the negative representation to eliminate the correlation between different classes such that the incoherence criterion of the extra-class representation is boosted. Moreover, a constructed subspace model is developed to enhance the self-expressive power of training samples and further build the representation bridge between the training and test samples, such that the coherence of the learned intra-class representation is consistently heightened. Finally, the resulting optimization problem is solved elegantly by employing an alternative optimization strategy, and a simple recognition algorithm on the learned representation is utilized for final prediction. Extensive experimental results demonstrate that the proposed method achieves superb recognition results on four face image datasets, three character datasets, and the fifteen scene multi-categories dataset. It not only shows superior potential on image recognition but also outperforms state-of-the-art methods.
\end{abstract}


\IEEEpeerreviewmaketitle

\section{Introduction}
\IEEEPARstart{D}{iscriminative} and effective data representations play an indispensable role in computer vision and machine learning, because they tremendously influence the performance of various learning systems \cite{Bengio2013}. A favorable data representation can greatly uncover the underlying information of observed data and intensely facilitate the machine learning methods \cite{ZZhang2015}. As a typical data representation method, sparse representation has earned its high reputation in both theoretical research and practical applications \cite{SRC, ZZhang2015, LLC2010}. Recently, low-rank representation (LRR) has captured considerable attention \cite{SLiTKDE, LRR, LatLRR}, and has also been proved to be a powerful solution to a wide range of applications, especially in subspace segmentation \cite{LRR}, feature extraction \cite{LatLRR} and image classification \cite{SLi2016, SXiao2016,YZhang2013}. In this paper, we focus on learning an appropriate data representation by constructing a block-diagonal low-rank representation for image recognition.

Sparse representation has been widely studied and applied in signal processing, machine learning and computer vision \cite{SRC,MElad2010}. The key idea of sparse representation is based on the assumption that each signal can be approximately represented by a linear combination of a few atoms of an over-completed dictionary. With the successful application of sparse representation based classification (SRC) \cite{SRC} in face recognition, numerous SRC based modifications have been proposed. For example, Nie et al. introduced an efficient and robust feature selection method by imposing the $l_{21}$-norm constraint on both loss function and regularization terms \cite{FNie2010}. Xu et al. \cite{YXu2011} proposed the semi-supervised sparse representation by employing a coarse-to-fine strategy, and Lu et al. \cite{CLu2013} developed a weighted sparse representation based classifier by leveraging both data locality and linearity to sparse coding. Based on the basic theorem \cite{LLC2010} that locality can always lead to sparsity but not necessarily vice versa, the locality-constrained linear coding (LLC) \cite{LLC2010} method achieves the sparse target by enforcing the locality embedding of codebook. In addition, some researchers argue that sparsity is not the ultimate reason of achieving decent recognition results \cite{CRC, LRC, ZZhangNC, YXu2016}. For example, Zhang et al. \cite{CRC} presented a collaborative representation based classification (CRC) method by employing the $l_2$-norm regularization rather than $l_1$-norm regularization for face recognition. It is demonstrated that CRC can achieve comparable performance but more efficient than SRC \cite{CRC}. The linear regression based classification (LRC) \cite{LRC} is another well-known representation based method. More specifically, LRC exploits each class of training samples to represent the test sample, and classifies it to the class leading to the minimum representation residual. A recent survey \cite{ZZhang2015} comprehensively reviews most representative sparse representation based algorithms, and empirically summarizes its wide applications from both theoretical and practical perspectives.

Recently, low-rank representation has gained increasing interest from different research fields. It is noted that the sparsity constraint can only dominate the local structure of each data vector, whereas the low-rank constraint can directly control the global structure of data \cite{LZhuang2012}. Furthermore, low-rank representation can greatly capture the underlying correlation behind the observed data \cite{LZhuang2012,XCai2013}. The most representative low-rank method, robust principal component analysis (RPCA) \cite{RPCA}, was proposed to recover the clean data with the low-rank constraint from corrupted observations. In particular, RPCA first assumes that the observations lie in a single subspace such that they can be decomposed into two separate components, i.e. the low-rank and sparse noise parts. However, RPCA cannot handle the situation where corrupted or outlying data are drawn from a union of multiple subspaces. To this end, Liu et al. \cite{LRR} proposed to perform matrix recovery by exploiting low-rank representation for subspace segmentation. The latent low-rank representation (LatLRR) \cite{LatLRR} was then developed for joint subspace segmentation and feature extraction by discovering the hidden information underlying the observations. Moreover, lots of low-rank representation based dictionary learning methods were proposed for robust image classification. For example, Zhang et al. \cite{YZhang2013} constructed a structured low-rank representation (SLRR) by regularizing all training images of the same class to have the same representation code. However, the ideal structure in SLRR is questionable, because, though data from the same class usually lie in the same subspace, it does not mean that images belonging to the same class should have the same data representation. Wei et al. \cite{LRSI} developed a low-rank matrix approximation method by learning sub-dictionaries independently for each class and meanwhile enforcing the structural incoherence between different classes. Li et al. \cite{YLi2014} explored a class-wise block-diagonal structure (CBDS) dictionary learning method, which learned discriminative low-rank representation by imposing the class-wise structure constraint. In addition, some variations of low-rank representation based methods were proposed to solve different problems. For example, Li et al. \cite{SLi2015} designed a cross-view low-rank analysis framework to address the multi-view outlier detection problem. Zhuang et al. \cite{NNLRS} presented a nonnegative low-rank sparse graph construction method for semi-supervised learning. As a result, it is widely agreed that the low-rank criterion indeed can disclose the potential data structures of different classes or tasks' correlation patterns, such that the effectiveness of the learned data representation can be enhanced.

There is a well-attested fact that the sparse representation in \cite{SRC} is a discriminative representation, whereas it only considers the data representation of each input signal independently, which does not take advantage of the global structural information in the set. In addition, existing research \cite{LRSI} \cite{SLi2015, YLi2014} has demonstrated that imposing specific structures on the low-rank representation matrix is beneficial to improve the discriminative capability of data representation. However, the performance of these methods is still far from being satisfactory. The main reason may be that these methods cannot perfectly transfer the original data features to the discriminative feature representations. Based on the well-explored self-expression property \cite{SSC}, the ideal block-diagonal representation can capture the underlying data information of samples by embedding the global semantic structure information and discriminative identification capability \cite{YZhang2013}. Consequently, promising results can be achieved if the discriminative data representation with the block-diagonal structure is exploited for recognition. In this paper, a novel block-diagonal low-rank representation (BDLRR) method is proposed to learn discriminative data representations which can simultaneously shrink the off-block-diagonal components and highlight the block-diagonal representation under the framework of low-rank representation. More specifically, BDLRR first eliminates the negative representation and boosts the incoherence of the extra-class representation by minimizing the off-block-diagonal representation, such that it can remove the noisy representation and transfer the positive representation to the block-diagonal components. Furthermore, BDLRR constructs a subspace model to enhance the self-expressive power of training samples and simultaneously bridge the representation gap between the training and test samples in a semi-supervised manner, such that the coherence of the intra-class representation is further improved and the learned representations are consistent. Finally, we introduce an effective iterative algorithm to solve the resulting optimization problem, and our method is evaluated to verify its adaptive capabilities for different recognition tasks. In summary, our key contributions are summarized as follows:

(1) A discriminative block-diagonal data representation structure is designed to boost the incoherent power of the extra-class representation by jointly removing the negative representation from the off-block-diagonal components and conveying the positive representation to the block-diagonal structure, such that better discriminative data representations are obtained for recognition tasks.

(2) A constructed subspace structure is developed to enhance the coherence of the intra-class representation by simultaneously improving the self-expressive capabilities of training samples and further narrowing the representation gap between training and test samples. Moreover, a low-rank criterion is enforced to capture the underlying feature structures of different classes or tasks' correlation patterns such that more competent representation results are achieved.

(3) By virtue of the semi-supervised learning superiority, the well-defined representation learning framework simultaneously learns both of discriminative training and test representations to keep consistence of the learned representations for recognition. To accommodate our method for large-scale problems, the out-of-sample extension is further explored to deal with new data instances.

(4) An effective optimization strategy based on the alternating direction method of multipliers (ADMM) is developed to solve the resulting optimization problem, and the convergence analysis of the designed optimization problem is presented from both theoretical and experimental perspectives.

The rest of this paper is organized as follows. We briefly review the related work on the low-rank theory in Section \ref{RelWork}. Then, we elaborate the proposed BDLRR method in Section \ref{proMethod}, and the solution to the optimization problem of the proposed BDLRR method is presented in Section \ref{optim}. Section \ref{Exp} reports extensive experimental results, as well as convergence and parameter sensitiveness analysis. Finally, the conclusion remarks are given in Section \ref{conc}.

\section{Related Work} \label{RelWork}
In this section, we give a brief review of two typical low-rank criterion based methods, i.e. robust principal component analysis (RPCA) \cite{RPCA} and low-rank representation (LRR) \cite{LRR}.

Let us first introduce our notations used in this paper. Matrices are represented with bold uppercase letters, e.g. $\bm{X}$, and column vectors are denoted by bold lower letters, e.g. $\bm{x}$. In particular, $\bm{I}$ denotes an identity matrix, and the entries of a matrix or vector are denoted by using $[\cdot]$ with subscripts. The $i$-th row and $j$-th column element of matrix $\bm{X}$ is denoted as $\bm{x}_{ij}$, and the block-diagonal matrix composed of a collection of matrices $[\bm{X}_1,\dots,\bm{X}_C]$ is denoted by
\begin{equation*}
diag(\bm{X}_1,\dots,\bm{X}_C) =
\left [
\begin{array}{cccc}
\bm{X}_1 & \ldots      & 0 \\
    \vdots    & \ddots & \vdots \\
    0    & \ldots & \bm{X}_C\\
\end{array}
\right ].
\end{equation*}

A matrix's $l_0$, $l_1$ and $ l_{21}$ norms are denoted as $\|\bm{X}\|_0=\sharp\{(i,j):\bm{x}_{ij} \neq 0\}$, $\|\bm{X}\|_1=\sum_{ij}|\bm{x}_{ij}|$, and $\|\bm{X}\|_{21}=\sum_j || \bm{X}_{.j} ||$, respectively. The norm induced by the $l_{\infty}$-norm on the matrix is denoted as $||\bm{X}||_{\infty}=\max_i \sum_j |\bm{x}_{ij}|$.  The matrix Frobenius norm designates $\|\bm{X}\|_F^2=tr(\bm{X}^T\bm{X})=tr(\bm{XX}^T)=\sum_{ij}\bm{x}_{ij}^2$, where $tr(\bullet)$ is the trace operator. $\|\bm{X}\|_*$ is the trace or nuclear norm of matrix $\bm{X}$, i.e. $\|\bm{X}\|_*=\sum_i |\sigma_i|$, where $\sigma_i$ is the $i$-th singular value of matrix $\bm{X}$. $\bm{X}^T$ denotes the transposed matrix of $\bm{X}$. $\mathbf{0}_{mn}$ denotes an all-zero matrix with the size of $m \times n$, and the all-one vector $\mathbf{1}_{N}=[\underbrace{1,\cdots,1}_N]^T$.

\subsection{Robust principal component analysis (RPCA)}
Suppose that $\bm{X}=[\bm{x}_1,\cdots,\bm{x}_n] \in \Re ^{d\times n}$ is the observed data matrix and composed of $n$ samples, where each column is a sample vector and usually has been highly corrupted. The main objective of RPCA is to determine a low-rank matrix $\bm{X}_0$ from the corrupted observations $\bm{X}$, and meanwhile filter out the sparse noise components $\bm{E}$, i.e. $\bm{X}=\bm{X}_0+\bm{E}$. Consequently, the objective function of RPCA can be easily formulated as
\begin{equation} \label{eq_1}
\min_{\bm{X}_0,\bm{E}}~rank(\bm{X}_0)+ \lambda ||\bm{E}||_0~~s.t.~~\bm{X}=\bm{X}_0+\bm{E},
\end{equation}
where the $rank(\bullet)$ operator denotes the rank of matrix $\bm{X}_0$, $\lambda$ is the balance parameter, and $\|\cdot\|_0$ means the $l_0$ pseudo-norm. Given an appropriate value of $\lambda$, RPCA can recover the clean data by $\bm{X}_0$. Due to the discrete properties of the rank function and the $l_0$-norm minimization, both of them are NP-hard problems and even difficult to approximate. An advisable choice \cite{RPCA} is to replace the rank constraint and $l_0$-norm regularization by the nuclear norm and $l_1$-norm regularization, respectively. As a result, problem (\ref{eq_1}) can be reformulated as
\begin{equation} \label{eq_2}
\min_{\bm{X}_0,\bm{E}}~||\bm{X}_0||_*+ \lambda ||\bm{E}||_1~~s.t.~~\bm{X}=\bm{X}_0+\bm{E},
\end{equation}
where $||\cdot||_*$ and $||\cdot||_1$ are the nuclear norm and $l_1$-norm, respectively. It is known that problem (\ref{eq_2}) can be efficiently solved by Augmented Lagrange Multiplier (ALM) method \cite{ZLin2010}.

\subsection{Low-rank representation (LRR) based method}
It is noted that RPCA is essentially based on the priori hypothesis that the observed data is approximately drawn from a low-rank subspace, that is, data can be described by a single subspace \cite{LRR}. However, this assumption is very difficult to be satisfied for real-world datasets, where multiple subspaces are more reasonable. To this end, LRR \cite{LRR} assumes that each data can be approximately represented by a union of several linear low-rank subspaces. The objective function of LRR is formulated as
\begin{equation} \label{eq_3}
\min_{\bm{Z},\bm{E}}~rank(\bm{Z})+ \lambda ||\bm{E}||_l~~s.t.~~\bm{X}=\bm{DZ}+\bm{E},
\end{equation}
where $\bm{D}$ and $\lambda$ are the dictionary and balance parameter, respectively. $||\cdot||_l$ indicates the constraint of different norms, and imposing different norms tends to remove specific noise as illustrated in \cite{LRR}. For example, the matrix Frobenius norm can effectively capture Gaussian noise, while the $l_1$-norm can better process the random noise or corruptions. Similar to RPCA, problem (\ref{eq_3}) can be approximately reformulated as
\begin{equation} \label{eq_4}
\min_{\bm{Z},\bm{E}}~||\bm{Z}||_*+ \lambda ||\bm{E}||_l~~s.t.~~\bm{X}=\bm{DZ}+\bm{E},
\end{equation}
which can also be effectively solved by the ALM algorithm \cite{LRR, ZLin2010}.

\section{The proposed Block-Diagonal Low-Rank Representation} \label{proMethod}
In this section, we introduce a novel block-diagonal low-rank representation (BDLRR) method, which collaboratively learns appropriate block-diagonal representations of training and test samples by jointly enforcing the incoherence of extra-class data representations and enhancing the coherence of intra-class data representations.

\textbf{Assumption 1}: \textsl{Let $\bm{X} = [\bm{X}_1,\cdots,\bm{X}_C]\in \Re^{d\times n}$ denote $N$ training samples with a dimension of $d$ from $C$ classes, where each column is a sample vector. Suppose that all the samples are rearranged based on the class labels, and each class of training samples are stacked together to form a sub-matrix $\bm{X}_i \in \Re^{d\times n_i}$, which denotes $n_i$ samples from the $i$-th class (i=1,2, .., C).}

\textbf{\textbf{Definition 1} (self-expressiveness property)}\cite{SSC}: Each data instance from a union of multiple subspaces can be effectively represented by a linear combination of other data instances, which is referred as the \emph{self-expressiveness property}.

\textbf{\textbf{Definition 2}}: Suppose that a data point $\bm y \in \Re^d$ is from the $i$-th class. $\bm{z}$ is a solution of the linear equation $\bm{y} = \bm{Xz}$, where
the sub- vectors $\bm{z}_j$ (j=1,2, .., C) of $\bm{z}$ respectively corresponding to the $j$-th class.
Based on the self-expressiveness property, the sub-matrix $\bm{X}_i$ should be able to well represent $\bm y$, and there is $\bm{y}\approx\bm{X}_i\bm{z}_i$. We define $\bm{z}_i$ as the \emph{intra-class representation}, otherwise the coding coefficients $\bm{z}_j$($j\neq i$) are called the \emph{extra-class representation}.

It is worth noting that the self-expressiveness property has already been successfully utilized in the context of classification \cite{SRC, LRSI}, low-rank matrix approximation \cite{LRR} and clustering \cite{SSC}. Typically, SSC and LRR are the most representative methods, and the explicit self-expressiveness formulation, $\bm{X}=\bm{XZ}$, is easily satisfied, where $\bm{Z}$ is data representation. Furthermore, in the presence of the self-expressiveness property, the key underlying observation of SSC and LRR is disclosed that each data point in a dataset can be ideally represented by a linear combination of a few points from its own subspace. Based on this observation and Assumption 1, the desired self-expressive representation should be block-diagonal and the obtained data representation is sufficiently discriminative. So, the ideal block-diagonal structure based representation is
\begin{equation} \label{eq_5}
\bm{X}=\bm{X\hat{Z}}~~s.t.~~\hat{\bm{Z}} = diag(\bm{Z}),
\end{equation}
where $\bm{Z} = [\bm{Z}_{11},\cdots,\bm{Z}_{CC}]$, and $\bm{Z}_{ij}$ is the representation coefficient of $\bm{X}_i$ corresponding to $\bm{X}_j$. However, the absolute block-diagonal structure is not easy to learn. To this end, it is natural to assume that the off-block-diagonal components are as small as possible to enhance the incoherent extra-class representation, which means that $\bm{Z}_{ij}$ tends to a zero sub-matrix for $i \neq j$. In addition, the coherent intra-class representation at the same time is further boosted. We formulate the following structured representation as
\begin{equation} \label{eq_6}
\min_{\bm{Z}} \lambda_1 ||\bm{A} \odot \bm{Z} ||_F^2 + \lambda_2 ||\bm{D} \odot \bm{Z} ||_0 ~~s.t.~~\bm{X} = \bm{XZ},
\end{equation}
where $\lambda_1$ and $\lambda_2$ are positive constants to weigh corresponding terms, $\odot$ indicates the Hadamard product (i.e. element-wise product), and $\bm{X} \in \Re^{d\times n}$. More specifically, the first term attempts to minimize the off-block-diagonal entries, and $\bm{A}=\bm{1}_n \bm{1}_n^T-\bm{Y}$ where $\bm{Y} =
\left [\begin{array}{cccc}
\bm{1}_{n_1} \bm{1}_{n_1}^T & \ldots      & 0 \\
    \vdots    & \ddots & \vdots \\
    0    & \ldots & \bm{1}_{n_C} \bm{1}_{n_C}^T\\
\end{array}\right ]$.
The second term is the constructed subspace measure to improve the coherent representation of intra-class representation. $\bm{d_{ij}}$ is a distance metric between $\bm{x}_i$ and $\bm{x}_j$ such that similar samples have high probabilities to be similar data representations. There are many distance metric methods. In this work, we simply define the distance between two samples as the square of the Euclidean distance, i.e. $||\bm{x}_i-\bm{x}_j||_2^2$. Because solving the $l_0$-norm minimization problem is an NP-hard problem, a relaxed counterpart of the second term is formulated as $||\bm{D} \odot \bm{Z} ||_1$. Thus, problem (\ref{eq_6}) can be reformulated as
\begin{equation} \label{eq_7}
\min_{\bm{Z}} \lambda_1 ||\bm{A} \odot \bm{Z} ||_F^2 + \lambda_2 ||\bm{D} \odot \bm{Z} ||_1 ~~s.t.~~\bm{X} = \bm{XZ}.
\end{equation}

In general, a low-rank criterion can further capture the underlying classes' correlation patterns such that the performance of resulting models can be improved \cite{XCai2013, LRR, YZhang2013}. By integrating problems (\ref{eq_7}) and (\ref{eq_4}), we propose the following objective function for the semi-supervised BDLRR:
\begin{equation} \label{eq_8}
\begin{split}
\min_{\bm{Z},\bm{E}} ||\bm{Z}||_*+ \lambda_1||\bm{\tilde{A}} \odot \bm{Z} ||_F^2 + \lambda_2 ||\bm{D} \odot \bm{Z} ||_1 \\ + \lambda_3 ||\bm{E}||_{21} ~~s.t.~~\bm{X} = \bm{X}_{tr}\bm{Z}+\bm{E},
\end{split}
\end{equation}
where $\lambda_1$, $\lambda_2$ and $\lambda_3$ are positive scalars that weigh the corresponding terms in (\ref{eq_8}). $\bm{X}_{tr}\in \Re^{d\times n}$ is the training data matrix and $\bm{X} \in \Re^{d \times N}$ includes both training and test samples, i.e. $\bm{X} = [\bm{X_{tr}},\bm{X_{tt}}]$. For the second term, $\bm{\tilde{A}}$ = $[\bm{A},\bm{1}_n \bm{1}_{N-n}^T]$ where $\bm{A}$ is the same as (\ref{eq_6}), and data representation $\bm{Z}=[\bm{Z}_{tr}, \bm{Z}_{tt}]$ such that an implicit $||\bm{Z}_{tt}||_F^2$ term is imposed to avoid overfitting. For the third term, $\bm{D} \in \Re^{n\times N}$ is denoted as the distance between training samples $\bm{X}_{tr}$ and all samples $\bm{X}$ such that the coherent representation of both $\bm{X}_{tr}$ and $\bm{X}_{tt}$ corresponding to $\bm{X}_{tr}$ can be enhanced simultaneously. $\bm{E}$ denotes the noise term with the $l_{21}$-norm regularization to capture sample-specific noise information \cite{LRR}. Moreover, data representation $\bm{Z}$ of training and test samples is incorporated into a unified optimization problem such that $\bm{Z}_{tr}$ and $\bm{Z}_{tt}$ are both optimal and discriminative.

\section{Optimization and algorithm analysis} \label{optim}
To solve the optimization problem of BDLRR in (\ref{eq_8}), we propose to utilize an alternating direction method, and separate the problem into several subproblems, which have close-form solutions.
\subsection{Optimization Algorithm}\label{optimAlg}
To solve optimization problem (\ref{eq_8}), we first make an equivalent transformation by introducing two auxiliary variables to make the problem separable, and then problem (\ref{eq_8}) can be rewritten as
\begin{equation} \label{eq_9}
\begin{split}
\min_{\bm{P,Z,Q,E}} ||\bm{P}||_*+ \frac{\lambda_1}{2} ||\bm{\tilde{A}} \odot \bm{Z} ||_F^2 + \lambda_2 ||\bm{D} \odot \bm{Q} ||_1 +\lambda_3 ||\bm{E}||_{21} \\
s.t.~~\bm{X} = \bm{X}_{tr}\bm{Z}+\bm{E},~\bm{P}=\bm{Z},~\bm{Q}=\bm{Z}.
\end{split}
\end{equation}

Then, we can get the following objective function of the problem by the augmented Lagrangian multiplier method. Here the augmented Lagrangian function of problem (\ref{eq_9}) is
\begin{equation} \label{eq_10}
\begin{aligned}
\mathcal L&(\bm{P,Z,Q,E,C_1,C_2,C_3})=||\bm{P}||_*+ \frac{\lambda_1}{2}||\bm{\tilde{A}} \odot \bm{Z} ||_F^2+ \\& \lambda_2 ||\bm{D} \odot \bm{Q} ||_1 +\lambda_3 ||\bm{E}||_{21}+ \langle \bm{C}_1, \bm{X}-\bm{X}_{tr}\bm{Z}-\bm{E} \rangle \\& + \langle\bm{C}_2, \bm{P}-\bm{Z} \rangle+ \langle \bm{C}_3, \bm{Q}-\bm{Z}\rangle + \frac{\mu}{2}(||\bm{P}-\bm{Z}||_F^2 + \\& ||\bm{X}-\bm{X}_{tr}\bm{Z}-\bm{E}||_F^2 + ||\bm{Q}-\bm{Z}||_F^2 ),\\
\end{aligned}
\end{equation}
where $\langle \bm{P},\bm{Q} \rangle = tr (\bm{P}^T\bm{Q})$. $\bm{C}_1$, $\bm{C}_2$ and $\bm{C}_3$ are the Lagrangian multipliers, and $\mu > 0$ is a penalty parameter. The augmented Lagrangian is minimized along one coordinate direction at each iteration, i.e. minimizing the loss with respect to one variable with the remaining variables fixed. We introduce the detailed procedures as follows.

\begin{table}[!t]
\centering
\begin{tabular}{m{75mm}}
\toprule
\textbf{Algorithm 1.} Solving Problem (\ref{eq_8}) by ADMM\\
\midrule
\textbf{Require:} All feature matrix $\bm{X}=[\bm{X}_{tr}, \bm{X}_{tt}]$; Parameters $\lambda_1$, $\lambda_2$, $\lambda_3$; Distance measure matrix $\bm{D}$.\\
\textbf{Initialization:}  $\bm{P}=\bm{0}$, $\bm{Z}=\bm{0}$, $\bm{Q}=\bm{0}$, $\bm{E}=\bm{0}$, $\lambda_1$, $\lambda_2$, $\lambda_3 > 0$, $\bm{C}_1=\bm{0}$, $\bm{C}_2=\bm{0}$, $\bm{C}_3=\bm{0}$, $\mu_{max}$ = $10^8$, \textsl{tol} = $10^{-6}$, $\rho$ = 1.15.\\

\textbf{While} not converged \textbf{do}\\
~~~~1).~Update $\bm{Z}$ by using (\ref{eq_16});\\
~~~~2).~Update $\bm{P}$ by using (\ref{eq_12});\\
~~~~3).~Update $\bm{Q}$ by using (\ref{eq_18});\\
~~~~4).~Update $\bm{E}$ by using (\ref{eq_21});\\
~~~~5).~Update Lagrange multipliers $\bm{C_1}$, $\bm{C_2}$ and $\bm{C_3}$:\\
~~~~~~~~$\left\{
\begin{array}{l}
\bm{C}_1^{t+1} = \bm{C}_1^t+\mu^t(\bm{X}-\bm{X}_{tr}\bm{Z}^{t+1}-\bm{E}^{t+1})\\
\bm{C}_2^{t+1} = \bm{C}_2^t+\mu^t(\bm{P}^{t+1}-\bm{Z}^{t+1}) \\
\bm{C}_3^{t+1} = \bm{C}_3^t+\mu^t(\bm{Q}^{t+1}-\bm{Z}^{t+1}). \\
\end{array}\right.$\\
~~~~6).~Update $\mu$:\\
~~~~~~~~~$\mu^{t+1}$ = min($\mu_{max}$, $\rho \mu^t$) \\
~~~~7).~Check convergence: if\\
~~~~~~~~~$\max \left( \begin{aligned}
&||\bm{X}-  \bm{X}_{tr}\bm{Z}^{t+1}-\bm{E}^{t+1}||_{\infty}, \\  ||\bm{P}^{t+1}& - \bm{Z}^{t+1}||_{\infty}, ||\bm{Q}^{t+1}-\bm{Z}^{t+1}||_{\infty}
 \end{aligned} \right) \leq$ \textsl{tol},\\~~~~~~~~and then stop.\\
\textbf{End While}\\
\bottomrule
\end{tabular}
\end{table}

\textsl{Updating $\bm{Z}$}:  Fix the other variables and update $\bm{Z}$ by solving the following problem
\begin{equation} \label{eq_14}
\begin{split}
\mathcal L &= \min_{\bm{Z}} \frac{\lambda_1}{2}||\bm{\tilde{A}} \odot \bm{Z} ||_F^2 + \langle \bm{C}_1^t, \bm{X}-\bm{X}_{tr}\bm{Z}-\bm{E}^t \rangle
 \\&~~+ \langle\bm{C}_2^t, \bm{P}^{t+1}-\bm{Z} \rangle + \langle \bm{C}_3^t, \bm{Q}^t-\bm{Z}\rangle  + \frac{\mu^t}{2}(||\bm{P}^{t+1}-\bm{Z}||_F^2 \\&~~~~~~~~~~~~~ +||\bm{X}-\bm{X}_{tr}\bm{Z}-\bm{E}^t||_F^2 + ||\bm{Q}^t-\bm{Z}||_F^2 )
\\& = \frac{\lambda_1}{2}||\bm{\tilde{A}} \odot \bm{Z} ||_F^2 + \frac{\mu^t}{2}(||\bm{X}-\bm{X}_{tr}\bm{Z}-\bm{E}^t+\frac{\bm{C}_1^t}{\mu^t}||_F^2 \\ & ~~~~~~~~~~ + ||\bm{P}^{t+1}-\bm{Z}+\frac{\bm{C}_2^t}{\mu^t}||_F^2+ ||\bm{Q}^t-\bm{Z}+\frac{\bm{C}_3^t}{\mu^t}||_F^2 ),
\end{split}
\end{equation}
which is equivalent to
\begin{equation} \label{eq_15}
\begin{aligned}
\mathcal L &=\min_{\bm{Z}} \frac{\lambda_1}{2}||\bm{Z} - \bm{R} ||_F^2 + \frac{\mu^t}{2}(||\bm{X}-\bm{X}_{tr}\bm{Z}-\bm{E}^t+\frac{\bm{C}_1^t}{\mu^t}||_F^2 \\ & ~~~~~~~~~~ + ||\bm{P}^{t+1}-\bm{Z}+\frac{\bm{C}_2^t}{\mu^t}||_F^2+ ||\bm{Q}^t-\bm{Z}+\frac{\bm{C}_3^t}{\mu^t}||_F^2 ),\end{aligned}
\end{equation}
where $\bm{R} = [\bm{Y},\bm{0}_{n(N-n)}]\odot \bm{Z}^t$. By setting the derivation $\frac{\partial \mathcal L}{\partial \bm{Z}} = \bm{0}$, we can easily infer the optimal solution of $\bm{Z}$, and the closed-form solution to problem (\ref{eq_15}) is given by the following form,
\begin{equation} \label{eq_16}
\begin{aligned}
\bm{Z}^{t+1} & = \left[ (2+\frac{\lambda_1}{\mu^t})\bm{I} + \bm{X}_{tr}^T\bm{X}_{tr} \right]^{-1} (\frac{\lambda_1}{\mu^t}\bm{R}+\bm{X}_{tr}^T \bm{S}_1+\bm{S}_2+\bm{S}_3),
\end{aligned}
\end{equation}
where $\bm{S}_1 = \bm{X}-\bm{E}^t + \frac{\bm{C}_1^t}{\mu^t}$, $\bm{S}_2 = \bm{P}^{t+1}+\frac{\bm{C}_2^t}{\mu^t}$, and $\bm{S}_3 = \bm{Q}^t+\frac{\bm{C}_3^t}{\mu^t}$.

\textsl{Updating $\bm{P}$}: When fixing the other variables, the objective function of (\ref{eq_10}) is degenerated into a function with respect to $\bm{P}$, i.e.
\begin{equation} \label{eq_11}
\begin{aligned}
\bm{P}^{t+1} & = \argmin_{\bm{P}}  ||\bm{P}||_*+ \langle\bm{C}_2^t, \bm{P}-\bm{Z}^t \rangle + \frac{\mu^t}{2}||\bm{P}-\bm{Z}^t||_F^2 \\
& = ||\bm{P}||_*+ \frac{\mu^t}{2}||\bm{P}- (\bm{Z}^t - \frac{\bm{C}_2^t}{\mu^t})||_F^2.
\end{aligned}
\end{equation}

This problem has a closed-form solution by using the singular value thresholding (SVT) operator \cite{ZLin2010} \cite{SVTM}, i.e.
\begin{equation} \label{eq_12}
\bm{P}^{t+1} = \mathcal T_{\frac{1}{\mu^t}}(\bm{Z}^t - \frac{\bm{C}_2^t}{\mu^t}) = \bm{U} \mathcal S_{\frac{1}{\mu^t}}(\bm{\Sigma}) \bm{V}^T,
\end{equation}
where $\bm{U \Sigma V}^T$ is the singular value decomposition of ($\bm{Z}^t - \frac{\bm{C}_2^t}{\mu^t}$), and $\mathcal S_{\frac{1}{\mu^t}} (\cdot) $ is the soft-thresholding operator \cite{ZZhang2015}\cite{ZLin2010}, which is defined as
\begin{eqnarray} \label{eq_13}
\mathcal S_{\lambda}(\bm{x}) =
\left\{
\begin{array}{ll}
\bm x - \lambda, ~~& if~~ \bm x > \lambda \\
\bm x + \lambda, ~~& if~~ \bm x < -\lambda \\
0, & otherwise.\\
\end{array}
\right.
\end{eqnarray}

\textsl{Updating $\bm{Q}$}: When the other variables are fixed, the objective optimization problem (\ref{eq_10}) with respective to $\bm{Q}$ is degenerated to the following problem

\begin{equation} \label{eq_17}
\begin{split}
\mathcal L = \min_{\bm{Q}} \lambda_2 ||\bm{D} \odot \bm{Q} ||_1 + \langle \bm{C}_3^t, \bm{Q}-\bm{Z}^{t+1}\rangle \\~~~~~~~~~~~~~+ \frac{\mu^t}{2}||\bm{Q}-\bm{Z}^{t+1}||_F^2\\
 =  \lambda_2 ||\bm{D} \odot \bm{Q} ||_1 + \frac{\mu^t}{2}||\bm{Q}-(\bm{Z}^{t+1}-\frac{\bm{C}_3^t}{\mu^t}) ||_F^2, \\
\end{split}
\end{equation}
which can be updated by the element-wise strategy. Obviously, problem (\ref{eq_17}) can be equivalently decoupled into $n \times N$ subproblems. For the $i$-th row and $j$-th column element $\bm{Q}_{ij}$, the optimal solution of problem (\ref{eq_17}) is
\begin{equation} \label{eq_18}
\begin{aligned}
 \bm{Q}_{ij}^{t+1}& = \argmin_{\bm{Q}_{ij}}~ \lambda_2 \bm{D}_{ij} |\bm{Q}_{ij}| + \frac{\mu^t}{2}(\bm{Q}_{ij}- \bm{M}_{ij})^2 \\
 & = \mathcal S_{\frac{\lambda_2 \bm{D}_{ij}}{\mu^t}} (\bm{M}_{ij}),
\end{aligned}
\end{equation}
where $\bm{M}_{ij} = \bm{Z}_{ij}^{t+1}-\frac{(\bm{C}_3^t)_{ij}}{\mu^t}$.

\textsl{Updating $\bm{E}$}: When other variables are fixed, problem (\ref{eq_10}) can be converted into the following problem
\begin{equation} \label{eq_19}
\begin{split}
 \min_{\bm{E}}\lambda_3 ||\bm{E}||_{21}+ \langle \bm{C}_1^t, \bm{X}-\bm{X}_{tr}\bm{Z}^{t+1}-\bm{E} \rangle \\ + \frac{\mu^t}{2} ||\bm{X}-\bm{X}_{tr}\bm{Z}^{t+1}-\bm{E}||_F^2,
\end{split}
\end{equation}
which is equivalent to
\begin{equation} \label{eq_20}
 \min_{\bm{E}}\lambda_3 ||\bm{E}||_{21}+ \frac{\mu^t}{2} ||\bm{E}-(\bm{X}-\bm{X}_{tr}\bm{Z}^{t+1}+ \frac{\bm{C}_1^t}{\mu^t})||_F^2.
\end{equation}

The solution to problem (\ref{eq_20}) is demonstrated in \cite{FNie2010}. In particular, let $\bm{\Gamma} = \bm{X}-\bm{X}_{tr}\bm{Z}^{t+1}+ \frac{\bm{C}_1^t}{\mu^t}$, the $i$-th row of $\bm{E}^{t+1}$ is
\begin{eqnarray}  \label{eq_21}
\bm{E}^{t+1}(i,:) = \left\{
\begin{array}{ll}
\frac{||\bm{\Gamma}^i||_2-\frac{\lambda_3}{\mu^t}}{||\bm{\Gamma}^i||_2} \bm{\Gamma}^i, ~~& if~~ ||\bm{\Gamma}^i||_2 > \frac{\lambda_3}{\mu^t} \\
0, ~~& if~~ ||\bm{\Gamma}^i||_2 \leq \frac{\lambda_3}{\mu^t}, \\
\end{array}\right.
\end{eqnarray}
where $\bm{\Gamma}^i$ is the $i$-th row of matrix $\bm{\Gamma}$. Here we denote the solution of $\bm{E}$ as $\mathcal H_{\frac{\lambda_3}{\mu^t}}(\bm{\Gamma})$ for convenience.

After we optimize variables $\bm{P}$, $\bm{Z}$, $\bm{Q}$ and $\bm{E}$, the ADMM algorithm also needs to update the Lagrange multipliers $\bm{C}_1$, $\bm{C}_2$, $\bm{C}_3$, as well as parameter $\mu$, for faster convergence. The detailed procedures of solving the proposed optimization problem (\ref{eq_8}) are described in Algorithm 1.

\begin{table}[!t]
\centering
\begin{tabular}{m{77mm}}
\toprule
\textbf{Algorithm 2.} BDLRR model for recognition\\
\midrule
\textbf{Input:} Training feature set $\bm{X}_{tr}$ with label matrix $\bm{Y}$, test sample set $\bm{X}_{tt}$.\\
\textbf{Output:} Predicted label matrix $\bm{L}$ for test samples.\\
~~1).~Normalize all the samples of both training and test samples
\\~~~~~~to unit-norm by using $\bm{x}_i=\bm{x}_i/\|\bm{x}_i\|_2$.\\
~~2).~Exploit \textbf{Algorithm 1} to solve problem (\ref{eq_8}), and a discriminative
\\~~~~~representation matrix $\bm{Z}=[\bm{Z}_{tr},\bm{Z}_{tt}]$ is obtained. \\
~~3).~Employ Eqn. (\ref{eq_23}) to learn an optimal linear classifier $\bm{\hat{W}}$.\\
~~4).~Predict the label matrix $\bm{L}$ of test samples $\bm{X}_{tt}$ by utilizing
\\~~~~~~Eqn. (\ref{eq_24}) one by one.\\
\bottomrule
\end{tabular}
\end{table}

\subsection{Recognition Method}
When problem (\ref{eq_8}) is optimized by exploiting Algorithm 1, the discriminant data representation $\bm{Z}=[\bm{Z}_{tr}, \bm{Z}_{tt}]$ is obtained. We directly employ a simple linear classifier to perform final recognition \cite{YZhang2013}. A linear classifier $\bm{W}$ is learned based on the training data representation $\bm{Z}_{tr}$ and its corresponding label matrix $\bm{L} \in \Re^{C\times n}$ of training samples. The following optimization problem is considered
\begin{equation} \label{eq_22}
 \bm{\hat{W}} = \argmin_{\bm{W}} ||\bm{L} - \bm{W}\bm{Z}_{tr}||_F^2+ \gamma ||\bm{W}||_F^2,
\end{equation}
where $\gamma$ is a positive regularization parameter. It is easy to verify that problem (\ref{eq_22}) has the close-form solution, i.e.
\begin{equation} \label{eq_23}
\bm{\hat{W}}=\bm{L}\bm{Z}_{tr}^T(\bm{Z}_{tr} \bm{Z}_{tr}^T+ \gamma \bm{I})^{-1}.
\end{equation}

The identity of test sample $\bm{y}$, the $i$-th sample from the test dataset, is determined by judging
\begin{equation} \label{eq_24}
label(\bm{y}) = \arg \max_j (\bm{\hat{W}}\bm{z}^i),
\end{equation}
where $\bm{z}^i$ is the $i$-th column of matrix $\bm{Z}_{tt}$. The complete procedures of our BDLRR model for recognition are summarized in Algorithm 2.

\subsection{Convergence Analysis}\label{convProof}
To solve the proposed formulation (\ref{eq_8}), an iterative update scheme, the ADMM algorithm, is developed as shown in Section \ref{optimAlg}. This section presents a theoretical convergence proof of the proposed Algorithm 1.

\textbf{\emph{Proposition 1}}: \textit{Algorithm 1 is equivalent to a two-block ADMM.}

The classical ADMM is intended to solve problems in the form
\begin{equation}\label{eq_CV1}
\min_{\bm{z}\in \Re^n, \bm w\in \Re^m}f(\bm z)+h(\bm w)~s.t.~\bm{R z}+\bm{T w}=\bm{u},
\end{equation}
where $\bm R \in \Re^{p\times n}$, $\bm T \in \Re^{p\times m}$, $\bm u \in \Re^{p}$ and $f$ and $h$ are convex functions. It is apparent that ADMM for problem (\ref{eq_CV1}) can be directly extended to solve the matrix optimization problem as follows:
\begin{equation}\label{eq_CV2}
\min_{\bm{Z}\in \Re^{n\times N}, \bm W\in \Re^{m\times N}}f(\bm Z)+h(\bm W)~s.t.~\bm{RZ}+\bm{TW}=\bm{U},
\end{equation}
where $\bm U \in \Re^{p\times N}$. The augmented Lagrangian of problem (\ref{eq_CV2}), in the method of multipliers, is formulated as
\begin{equation}\label{eq_CV3}
\begin{aligned}
\mathcal L_{\mu}(\bm Z, \bm W, \bm C) = f(\bm Z)+&h(\bm W)\\+\frac{\mu}{2}\|\bm{RZ}+\bm{TW}-\bm{U}&\|_F^2+\langle \bm C,\bm{RZ}+\bm{TW}-\bm{U} \rangle,
\end{aligned}
\end{equation}
where $\bm C \in \Re^{p\times N}$ is the Lagrangian multiplier, and $\mu$ is a penalty coefficient.

It should be noted that problem (\ref{eq_9}) is a special case of problem (\ref{eq_CV2}). Specifically, it can be verified that the constraints in (\ref{eq_9}) can be transformed into the form of $\bm R\bm Z+\bm T \bm W=\bm U$, where $\bm{R} = \left( \begin{array}{c} -\bm I_n\\ -\bm I_n\\ \bm X_{tr}\\ \end{array} \right)$, $\bm T = \left [\begin{array}{ccc} \bm I_n & & \\ & \bm I_n & \\ & &\bm I_d\\ \end{array}\right]$, $\bm W = \left( \begin{array}{c} \bm P\\ \bm Q\\ \bm E\\ \end{array} \right)$, $\bm U = \left( \begin{array}{c} \bm 0\\ \bm 0\\ \bm X\\ \end{array} \right)$, and $\bm{I}_n$ is an $n\times n$ identity matrix. In this way, problem (\ref{eq_9}) is reformulated as problem (\ref{eq_CV2}). Moreover, ADMM updates two primal variables in an alternating fashion, and iteratively solves problem (\ref{eq_CV3}) as follows:
\begin{subequations}\label{eq_CV4}
	\begin{align}
        \bm{Z}^{t+1} &= \arg\min_{\bm{Z}\in \Re^{n\times N}}\mathcal L_{\mu}(\bm Z,\bm W^t, \bm C^t),\\
        \bm{W}^{t+1} &= \arg\min_{\bm{W}\in \Re^{m\times N}}\mathcal L_{\mu}(\bm Z^{t+1},\bm W, \bm C^t),\\
        \bm{C}^{t+1} &=  \bm C^t+\mu(\bm{RZ}^{t+1}+\bm{TW}^{t+1}-\bm{U}),
	\end{align}
\end{subequations}
which have the same updating procedures as Algorithm 1 in subsection \ref{optimAlg}. In fact, we can see that optimization of $\bm Z$ in (\ref{eq_CV4}a) is equivalent to optimize $\bm Z$ in (\ref{eq_14}). Furthermore, it is very important that when fixing $\bm Z$, solutions of $\bm P$ in (\ref{eq_12}), $\bm Q$ in (\ref{eq_18}), and $\bm E$ in (\ref{eq_21}), are independent on one another, for instance, computation of $\bm E^{t+1}$ only depends on $\bm Z^{k+1}$ and $\bm C^{k+1}$ rather than $\bm P^{k+1}$ or $\bm Q^{k+1}$. Hence, optimizations of $\bm P$, $\bm Q$ and $\bm E$ can be accumulated in $\bm W$ by using Eqn. (\ref{eq_CV4}b), updating of which is the same as fashion of Jacobian iterative method. In this way, problem (\ref{eq_9}) is a special case of classical ADMM problem (\ref{eq_CV2}), and the proposed optimization algorithm shown in Algorithm 1 has the same optimization style of classical ADMM (\ref{eq_CV4}). Therefore, the proposed optimization algorithm shown in Algorithm 1 is equivalent to a two-block ADMM, the global convergence of which is theoretically guaranteed \cite{Glowinski1989,Esser2009,Eckstein1992}. The convergence nature of the proposed optimization algorithm is given by the following theorem.
\begin{MyTheorem}\label{Theorem1}
(\cite{Esser2009,Eckstein1992}) Consider the problem (\ref{eq_CV1}) where $f(\bm Z)$ and $h(\bm W)$ are closed proper convex functions, $\bm R$ has full column rank and $h(\bm W)+\|\bm{TW}\|_F^2$ is strictly convex. Let $\bm C^0$ and $\bm W^0$ be arbitrary matrix and $\mu >0$. Assume that we have the sequences $\{\gamma_t\}$ and $\{\nu_t\}$ such that $\gamma_t\geq0$ and $ \nu_t\geq 0$, $\sum_{t=0}^{\infty}\gamma_t < \infty$ and $\sum_{t=0}^{\infty}\nu_t < \infty$. Suppose that
\begin{equation}\label{eq_CV5a}
\begin{split}
\|\bm{Z}^{t+1} - \min_{\bm{Z}}f(\bm Z) +\frac{\mu}{2}\|\bm{RZ}+\bm{TW}^t-\bm{U}\|_F^2\\+\langle \bm C^t,\bm{RZ}\rangle \|_F^2 \leq \gamma_t.
\end{split}
\end{equation}
\begin{equation}\label{eq_CV5b}
\begin{split}
\|\bm{W}^{t+1} - \min_{\bm{W}}h(\bm W) +\frac{\mu}{2}\|\bm{RZ}^{t+1}+\bm{TW}-\bm{U}\|_F^2\\+\langle \bm C^t,\bm{TW}\rangle\|_F^2 \leq \nu_t.
\end{split}
\end{equation}
\begin{equation}\label{eq_CV5c}
\begin{split}
 \bm{C}^{t+1} &=  \bm C^t+\mu(\bm{RZ}^{t+1}+\bm{TW}^{t+1}-\bm{U}).
\end{split}
\end{equation}
If there exists a saddle point of $\mathcal L_{\mu}(\bm Z,\bm W, \bm C)$ (\ref{eq_CV3}), then $\bm Z^k\rightarrow \bm Z^*$, $\bm W^k\rightarrow \bm W^*$ and $\bm C^k\rightarrow \bm C^*$, where ($\bm Z^*, \bm W^*, \bm C^* $) is such a saddle point. On the other hand, if no such saddle point exists, then at least one of the sequences $\{\gamma_t\}$ or $\{\nu_t\}$ must be unbounded.
\end{MyTheorem}

Clearly, the optimization results shown in subsection \ref{optimAlg} indicate that the proposed method exists an optimal solution according to the \textit{Proposition} $1.1.5$ in \cite{Bertsekas2009}, and the values sequences $\{\gamma_t\}$ and $\{\nu_t\}$ are directly set to zeros in Algorithm 1. Therefore, the convergence nature of our optimization method is demonstrated. Moreover, we empirically show in Section \ref{convergence} that the experimental convergence of the resulting ADMM is well preserved.

\subsection{Computational Complexity Analysis}
In this section, the computational complexity for Algorithm 1 is presented, and it is easy to see that the recognition process of Algorithm 2 is very efficient, which is linear with the sample number. More specifically, the major computation cost of Algorithm 1 is in steps 1-4, which require computing the singular value decomposition (SVD) and matrix computation operation. Thus, they will be time consuming when the number of training samples $n$ and the total number of samples $N$ are very large. In particular, computing SVD decomposition of matrix $\bm P \in \Re^{n\times N}$ needs the complexity of $\mathcal{O}(n^2N)$ $(N>n)$. Note that due to the matrix inverse calculation, calculating $\bm Z$ will scale in about $\mathcal{O}(n^2d+n^2N)$ where $d$ is the dimensionality of the samples. The computational complexity of step 3 is $\mathcal{O}(nN)$, and computing $\bm E$ in step 4 costs $\mathcal{O}(dN)$. Therefore, the total computational complexity of BDLRR is $\mathcal{O}_\kappa(2n^2N+n^2d+dN+nN)$, where $\kappa$ is the number of iterations.

In comparison, the computation burden of the sparse representation based classification methods such as SRC, LRSI and LatLRR are $\mathcal{O}(n^2(N-n)d)$ by solving ($N-n$) independent $l_1$-norm minimization problems in an iterative optimization manner \cite{SRC, CRC, LRSI}, which is slower than that of our method. The computation complexities of regression methods such as LRLR and LRRR are $\mathcal{O}(dn+n^2d)$, which is a little faster than our method. The low-rank and sparse representation based methods such as NNLRS, SRRS, CBDS, and our BDLRR need to simultaneously compute SVD of feature matrix and solve a simple soft-thresholding problem, and a linear classification algorithm is used to predict final labels of test data. Generally, the overall computation burden of our BDLRR is the same as those of the low-rank sparse representation learning methods.

\subsection{Out-of-sample Extension}\label{OE}
It is worth noting that the low-rank representation based methods have been extensively studied, but how to address the out-of-sample problem, the capability of dealing with new data instances, is much less well-solved. The stage of BDLRR mentioned above only obtains the discriminative representations of the available samples $\bm{X} \in \Re^{d\times N}$. However, given unseen instances outside the training and test data, it would be unrealistic and time-consuming to reimplement the whole model to produce the representations of novel images. In this subsection, we will show that the proposed BDLRR method can naturally cope with the out-of-sample examples to learn discriminative visual representations.

Suppose we have obtained the optimal block-diagonal representation $\bm{Z} \in \Re^{n\times N}$ from the available samples $\bm{X}$ over $\bm{X}_{tr} \in \Re^{d\times n}$ using the proposed model (\ref{eq_8}). Now, we extend the proposed BDLRR method to learn preferable representation of a novel image $\bm{b} \in \Re^{d\times 1}$ in the original observed space. Specifically, we aim at learning the discriminative representation $\bm z$ for $\bm{b}$ over $\bm{X}_{tr}$, while fixing the previously learned representation $\bm{Z}$. Therefore, adding terms for a novel data point $\bm b$ in model (\ref{eq_8}) and keeping the already learned variables, the objective function of the augmented BDLRR is formulated as
\begin{small}
\begin{equation} \label{eq_32}
\begin{split}
\min_{\bm{z},\bm{e}} ||[\bm{Z},\bm z]||_*+ \lambda_1||\bm{\hat{A}} \odot [\bm{Z},\bm z] ||_F^2 + \lambda_2 ||\bm{\hat{D}} \odot [\bm{Z},\bm z] ||_1 \\ + \lambda_3 ||[\bm{E},\bm e]||_{21} ~~s.t.~~[\bm{X},\bm b] = \bm{X}_{tr}[\bm{Z},\bm z]+[\bm{E},\bm e],
\end{split}
\end{equation}
\end{small}
where $\bm{\hat{A}}=[\bm{A},\bm{1}_n \bm{1}_{N+1-n}^T]$, $\bm{A}$ is defined as in Eqn. (\ref{eq_8}), $\bm{\hat{D}} \in \Re^{n\times (N+1)}$ is the distance metric between the training samples $\bm{X}_{tr}$ and all samples $[\bm{X},\bm b]$, and $\bm e$ is the representation error of $\bm b$ over $\bm{X}_{tr}$. We argue that $\|[\bm Z, \bm z]\|_* = \|\bm Z\|_*$. Particularly, for the learned representation $\bm Z \in \Re^{n\times N}$($n<N$), it is easy to find that the linear problem for $\bm\alpha$ in $\bm z = \bm {Z\alpha}$ is an underdetermined system for practical data. Generally speaking, $\bm z = \bm {Z\alpha}$ has infinitely many solutions in practice \cite{Trefethen1997}. Provided $n\ll N$, the matrix $\bm Z$ is row full rank and $\bm z = \bm {Z\alpha}$ has solution. In this way, the singular values of matrix $\bm Z$ coincide with those of $[\bm Z, \bm z]$, which means $rank([\bm Z, \bm z]) = rank(\bm Z)$. Therefore, $\|[\bm Z, \bm z]\|_* = \|\bm Z\|_*$ and it does not change for practical data in Eqn. (\ref{eq_32}). By removing the irrelevant terms with respective to the variables $\bm z$ and $\bm e$, it is easy to check that problem (\ref{eq_32}) will be degenerated to the following formulation:
\begin{equation} \label{eq_33}
\begin{split}
\min_{\bm{z},\bm{e}} \lambda_1\|\bm z\|_2^2 + \lambda_2 \|\bm{d} \odot \bm z \|_1+ \lambda_3 \|\bm e\|_2 ~~\\s.t.~~\bm b = \bm{X}_{tr}\bm z+\bm e,
\end{split}
\end{equation}
which can be equivalently reformulated as
\begin{equation} \label{eq_34}
\begin{split}
\min_{\bm{z},\bm{e}} \lambda_1\|\bm z\|_2^2 + \lambda_2 \|\bm{d} \odot \bm z \|_1+ \lambda_3 \|\bm e\|_2^2 ~~\\s.t.~~\bm b = \bm{X}_{tr}\bm z+\bm e,
\end{split}
\end{equation}
where $\bm d_{i}$ is the distance between $\bm x_i$ and $\bm b$. To make problem (\ref{eq_34}) more compact, it can be rewritten as
\begin{equation} \label{eq_35}
\begin{split}
\min_{\bm{z}} \frac{1}{2}\|\bm b - \bm{X}_{tr}\bm z\|_2^2 + \frac{\beta_1}{2}\|\bm z\|_2^2 + \beta_2 \|\bm{d} \odot \bm z \|_1,
\end{split}
\end{equation}
where $\beta_1=\lambda_1/\lambda_3$ and $\beta_2=\lambda_2/{2\lambda_3}$. Apparently, problem (\ref{eq_35}) is an elastic-net regularized regression problem. For convenient interpretation, we denote $g(\bm z) = \frac{1}{2}\|\bm b - \bm{X}_{tr}\bm z\|_2^2 + \frac{\beta_1}{2}\|\bm z\|_2^2$. With some algebra, problem (\ref{eq_35}) can be approximately transformed to the following optimization problem:
\begin{small}
\begin{equation} \label{eq_36}
\begin{aligned}
\bm z^{k+1} &= \arg\min_{\bm{z}} \beta_2 \|\bm{d} \odot \bm z \|_1 + \langle\nabla_{\bm z} g(\bm z^k), \bm z -\bm z^k \rangle + \frac{\eta}{2}\|\bm z-\bm z^k\|_2^2\\
&=\arg\min_{\bm{z}} \beta_2 \|\bm{d} \odot \bm z \|_1 + \frac{\eta}{2}\|\bm z-\bm z^k + \nabla_{\bm z} g(\bm z^k)/ \eta\|_2^2+const,\\
\end{aligned}
\end{equation}
\end{small}
where $\bm z^k$ is the $k$-th iteration of $\bm z$, and $\eta$ = $\|\bm X_{tr}\|_F^2$ is a fixed step size in our paper. Similar to problem (\ref{eq_18}), the optimal solution of the $i$-th entry of $\bm z$ is calculated by using $\bm z_i^{k+1} = \mathcal S_{\frac{\beta_2 \bm{d}_i}{\eta}} ([\bm z^k - \nabla_{\bm z} g(\bm z^k)/ \eta]_i)$. After obtaining the optimal solution $\bm z$, we identify the new data instance $\bm{b}$ by employing Eqn. (\ref{eq_24}), i.e. $label(\bm{b}) = \arg \max_j (\bm{\hat{W}}\bm{z})$. The promising recognition results can be guaranteed based on the observation that the discriminative block-diagonal training representations are learned in the training stage. Therefore, based on the proposed BDLRR model, the problem of recognizing new instances outside the training and test samples is well addressed.

\subsection{Discussion}
As we know, BDLRR simultaneously takes advantages of supervised information, i.e. label information, and semi-supervised learning superiority, i.e. learning training and test representations in one formulation. Moreover, our method intrinsically inherits the superiorities of sparse, low-rank, structured and elastic-net representation learning techniques. This characteristic naturally differentiates it from previous works, yielding superior recognition results. In this section, we establish the relationships between the proposed BDLRR method and some related discriminative low-rank representation methods, such as the nonnegative low-rank representation sparse (NNLRS) method \cite{NNLRS}, the structured sparse and low-rank representation (SSLR) method \cite{YZhang2013}, and the very recently proposed supervised regularization based robust subspace (SRRS) method \cite{SLi2015}.

\subsubsection{Connection to the NNLRS method}
The NNLRS method focuses on constructing the informative graph by jointly considering the low-rank and sparse representation to capture the global and local structures of data, respectively. Specifically, the objective function of NNLRS is formulated as
\begin{equation}\label{eq_p1}
\begin{split}
\min_{\bm{Z,E}} \|\bm{Z}\|_* + \lambda_2\|\bm{Z}\|_1 + \lambda_3\|\bm{E}\|_{21}~\\s.t.~\bm{X_{tr}} = \bm{X_{tr}Z} + \bm{E},\bm{Z} \ge \bm{0}.
\end{split}
\end{equation}
The rationale of NNLRS is under the guidance of the observation that the sparse constraint ensures each sample connected to only few other samples resulting in sparse representation, while the low-rank constraint enforces the learned representation from the same class with high correlations. In other words, NNLRS is designed to capture the global structure of the training data using the low-rank property, and the locality information of each data vector is interpolated into NNLRS by introducing the sparse term. The following proposition shows the close relationship between the proposed BDLRR method and the LRR and NNLRS methods.

\textbf{\emph{Proposition 2}}: \textit{The proposed BDLRR method is a generalized but discriminative low-rank representation learning model, and both of LRR and NNLRS are the special cases of the proposed BDLRR method.}
\begin{proof}[\textbf{Proof}]
From the objective function of BDLRR Eqn. (8), if we set both balance parameters $\lambda_1=0$ and $\lambda_2=0$, it is easy to find that BDLRR will degenerate to LRR along with learning the training and test representations in a semi-supervised manner. Moreover, if the penalty parameter $\lambda_1=0$ and $\bm X = \bm X_{tr}$, BDLRR will be reformulated as a low-rank and weighted sparse representation learning model. As a result, BDLRR will degenerate to weighted NNLRS without considering the nonnegative constraint. Therefore, both of LRR and NNLRS are the special cases of the proposed BDLRR method.

More importantly, our BDLRR method jointly considers suppressing the unfavorable representations from off-block-diagonal components and highlighting the compact block-diagonal representations under the framework of the semi-supervised low-rank representation learning such that the margins between different classes are greatly enlarged and the intra-class compactness is also enhanced simultaneously. In this way, BDLRR takes the intra-class and inter-class visual correlations into consideration to concurrently learn both discriminative representations of training and test data in one unified learning paradigm. As a result, our method can be viewed as a generalized discriminative representation learning framework.

Therefore, our BDLRR method not only intrinsically generalizes the previous LRR and NNLRS models, but also extends the existing low-rank representation models to more robust and discriminative cases.
\end{proof}

\subsubsection{Comparison with the SSLR method}
SSLR first learns a structured low-rank sparse dictionary by imposing an ideal representation regularization term, and then a structured low-rank representation is achieved based on the learned dictionary. The objective function of SSLR is
\begin{equation}\label{eq_p2}
\begin{split}
\min_{\bm{Z,E,\Xi}} \|\bm{Z}\|_* + \lambda_1\|\bm{Z}\|_1 + \lambda_2\|\bm{E}\|_1+\lambda_3\|\bm Z- \bm Q\|_F^2~\\s.t.~\bm{X_{tr}} = \bm{\Xi}\bm{Z} + \bm{E},
\end{split}
\end{equation}
where $\bm \Xi$ is the learned dictionary. $\bm Q$ is the ideal data representation of training samples, i.e. $\left [\begin{array}{cccc}
\bm{1}_{s_1} \bm{1}_{s_1}^T & \ldots      & 0 \\
    \vdots    & \ddots & \vdots \\
    0    & \ldots & \bm{1}_{s_C} \bm{1}_{s_C}^T\\
\end{array}\right ]$, where $s_i$ is the number of the $i$-th class of $\bm \Xi$. By solving the optimization problem (\ref{eq_p2}), the learned dictionary $\bm \Xi$ is obtained, and then representations of the training and test data are respectively achieved by directly removing the ideal representation term from (\ref{eq_p2}), resulting in the following optimization problem:
\begin{equation}\label{eq_p3}
\begin{split}
\min_{\bm{Z,E}} \|\bm{Z}\|_* + \lambda_1\|\bm{Z}\|_1 + \lambda_2\|\bm{E}\|_1~s.t.~\bm{B} = \bm{\Xi Z} + \bm{E},
\end{split}
\end{equation}
where $\bm B$ is the observations, i.e. $\bm X_{tr}$ or $\bm X_{tt}$. Although the experimental results reported in \cite{YZhang2013} are good, we hold the view that enforcing the representation approximal to the ideal representation matrix $\bm Q$ is questionable because it is impossible to regularize all the training samples of the same class to have the same representation codes. Moreover, the solution of learning $\bm \Xi$ by solving problem (\ref{eq_p2}) sensitively depends on the initialization because of the nonconvex optimization. Furthermore, learning representations of the training and test data are divided into two separate stages, and there are respectively three and two parameters in (\ref{eq_p2}) and (\ref{eq_p3}), which are very difficult to tune.

In contrast, our method is reasonable and discriminative. BDLRR first shrinks the off-block-diagonal elements to eliminate the unfavorable representations resulting in marginalized inter-class representations and highlights the block-diagonal elements yielding compact intra-class representations. In this way, the discriminative constraints in BDLRR simultaneously separates the common visual representations from different classes, and effectively prevents zero entities from appearing in the class-specific representations. Moreover, we believe that it is significant for recognition that the learned representations of training and testing samples should be consistent. To this end, BDLRR builds the representation bridge between the training and test samples by imposing the low-rank and locality coherence property. Thus, the proposed BDLRR method unifies the discriminative representations of training and test data into one robust learning framework such that better recognition results are achieved.

\subsubsection{Comparison with the SRRS method}
The main objective of SRRS is dedicated to learning a discriminative subspace from the clean data recovered by using the low-rank representation constraint. The main idea of SRRS is to remove noise from contaminated data depending on the denoising capability of the low-rank representation, and then the discriminative subspace is learned based on the recovered `clean' data. The objective function of SRRS is
\begin{equation}\label{eq_p4}
\begin{aligned}
\min_{\bm{Z,E,P}} &\|\bm{Z}\|_*+\lambda_2\|\bm{E}\|_{21} +\eta \|\bm P^T\bm{XZ}\|_F^2\\ &+\lambda_1 [tr(S_b(\bm P^T \bm{XZ}))-tr(S_w(\bm P^T \bm{XZ}))] \\~s.t.&~\bm{X}_{tr} = \bm{X}_{tr}\bm{Z} + \bm{E},\bm P^T \bm P =\bm I,
\end{aligned}
\end{equation}
where $S_b(\bullet)$ and $S_w(\bullet)$ are respectively the between-class and within-class scatter matrices, and $\eta$ is a balance parameter.

Apparently, our method is different from SRRS. First, SRRS directly utilizes the `clean' data $\bm{XZ}$ to perform discriminant analysis, which means that the performance of SRRS is greatly subject to the denoising ability of LRR. However, BDLRR aims at directly learning discriminative representations from data by imposing the discriminant constraints, which are not confined to any other conditions. Moreover, SRRS is a subspace learning method, and then the dimension selection of the final representations is very important for recognition. However, BDLRR directly learns discriminative representations from data, and recognition is performed on the optimal representations without bearing the burden of dimension selection. In addition, our BDLRR method jointly learns the representations of training and test data, whereas the test representations of SRRS is achieved by using $\bm{PX}_{tt}$, which can not capture the component connections between the learned representations of the training and test data. Therefore, the proposed BDLRR method is more robust and discriminative than SRRS, which is also verified by the subsequent experimental results.

\section{Experimental Validation} \label{Exp}
In this section, the performance of the proposed BDLRR method is evaluated for different recognition tasks. Extensive experiments are performed on different types of datasets to demonstrate the effectiveness and superiority of the proposed method in comparison with state-of-the-art recognition methods. Subsequently, the algorithmic convergence and the selection of parameters are well analyzed.

\subsection{Experimental Setup}
We test our method on eight benchmark datasets for three basic recognition tasks. Moreover, we compare with some state-of-the-art recognition methods, including representation based methods (such as LRC \cite{LRC}, CRC \cite{CRC}, SRC \cite{SRC} and LLC \cite{LLC2010}), low-rank criterion based methods (such as RPCA \cite{RPCA}, LatLRR \cite{LatLRR}, Low-rank linear regression (LRLR) \cite{XCai2013}, Low-rank robust regression (LRRR) \cite{XCai2013}, CBDS \cite{YLi2014}, LRSI \cite{LRSI}, NNLRS \cite{NNLRS}, SRRS \cite{SLi2015}), and conventional classification methods such as support vector machine (SVM) \cite{CChang2011} with Gaussian kernel. We randomly select several images per class to construct the training dataset, and the rest of images are regarded as the test set. All the selection processes are repeated 10 times, and the average recognition accuracies are reported for all the methods.


For fair comparison in all experiments, we use the Matlab codes from the corresponding authors with the default or optimal parameter settings, or directly cite the experimental results from their original papers. More specifically, for RPCA \cite{RPCA}, we first use the original RPCA algorithm on both training and test datasets to eliminate some noise and corrupted terms, and then exploit SRC \cite{SRC} for recognition. For LatLRR \cite{LatLRR}, the learned salient features are used for recognition. For SVM, the LibSVM software \cite{CChang2011} is used for multi-class recognition, where the important regularization parameter $C$ in SVM is selected by cross-validation from the candidate set \{0.01, 0.1, 1.0, 10.0, 100.0, 1000.0\}. The parameters of our method, i.e. $\lambda_1$, $\lambda_2$ and $\lambda_3$, are tuned to achieve the best performance via 5-fold cross validations from [0.1, 0.5, 1, 5, 10, 15, 20, 25]. To guarantee the same experimental settings between all the compared methods and our method on each benchmark, we re-implemented all the algorithms using respective optimal parameters via the cross-validation strategy, and the training and test samples were randomly selected from each dataset ten times. Since the scene character recognition datasets have the standard splits of the training and test data, we directly employ the full training and test data for recognition, and the compared experimental results are cited from the original papers. Similarly, for scene recognition, experiments are performed with the same experiment protocols as that of the LC-KSVD method \cite{ZJiang2013}, and we directly cite some experimental results from the original papers. For the compared methods that are not included in \cite{ZJiang2013}, we rerun them following the same experimental settings. Therefore, all the methods presented in our paper are performed on the same testbed for each dataset such that our experimental results are convincing and reliable. All algorithms are implemented with Matlab 2013a, and \textit{the Matlab code of the proposed method has been released at http://www.yongxu.org/lunwen.html}.

\begin{table}[!t]
\begin{center}
\caption{Recognition accuracies (mean$\pm$std \%) of different methods with different numbers of training samples on the Extended YaleB database.} \label{Table_1}
\begin{tabular}{|c|c|c|c|c|c|c|}
\hline
Alg. & 20 & 25 & 30 & 35\\
\hline \hline
LRC & 92.15$\pm$0.95 & 93.55$\pm$0.65 & 94.55$\pm$0.68 & 95.49$\pm$0.55 \\
CRC & 94.36$\pm$1.17 & 95.89$\pm$0.91 & 97.14$\pm$0.75 & 97.93$\pm$0.55 \\
SRC & 93.73$\pm$0.70 & 95.58$\pm$0.26 & 96.37$\pm$0.45 & 97.13$\pm$0.42 \\
LLC & 91.60$\pm$0.50 & 94.20$\pm$0.49 & 95.29$\pm$0.38 & 96.05$\pm$0.51 \\
SVM&92.81$\pm$0.68&95.20$\pm$0.44&96.11$\pm$0.41&96.70$\pm$0.69\\
RPCA&93.58$\pm$0.61&95.51$\pm$0.36&96.70$\pm$0.46&96.96$\pm$0.49\\
LatLRR&93.05$\pm$0.95&93.91$\pm$0.68&95.03$\pm$0.83&97.14$\pm$0.36\\
LRLR &83.91$\pm$1.53&85.15$\pm$1.50&85.49$\pm$1.05&85.95$\pm$1.47 \\
LRRR &83.95$\pm$0.82&85.66$\pm$0.93&86.21$\pm$0.99&86.55$\pm$0.81\\
CBDS&95.99$\pm$1.11&96.56$\pm$0.85&97.61$\pm$0.82&98.13$\pm$0.55\\
LRSI&94.19$\pm$0.44&96.28$\pm$0.61&96.99$\pm$0.57&97.72$\pm$0.48\\
NNLRS&94.35$\pm$0.79&96.06$\pm$0.63&97.02$\pm$0.61&97.62$\pm$0.42\\
SRRS&93.74 $\pm$0.86&96.05$\pm$0.95&96.89$\pm$0.84&97.15$\pm$0.58\\
\hline
\textbf{BDLRR}&\textbf{96.89}$\pm$\textbf{0.67}&\textbf{97.96}$\pm$\textbf{0.42}&\textbf{98.70}$\pm$\textbf{0.46}&\textbf{99.46}$\pm$\textbf{0.29}\\
\hline
\end{tabular}
\end{center}
\end{table}

\subsection{Experiments for Face Recognition}\label{ExFace}
In this section, we perform experiments on four face image datasets, including the Extended YaleB \cite{ExYaleB}, CMU PIE \cite{PIE}, AR \cite{AR} and LFW \cite{LFW} datasets.

\textbf{\emph{The Extended YaleB Database:}} The extended YaleB database is composed of 2414 face images of 38 subjects, where each person has 59-64 near frontal images under different illumination conditions. 
All the images for our experiments on this database have been resized to 32$\times$32 pixels. For all the compared methods, the suggested parameters from the corresponding papers are used for recognition. For the LLC \cite{LLC2010} method, we directly treat the training samples as the bases, and the coding coefficients are obtained using the approximated LLC strategy. The number of neighbors of LLC is set to fifteen for this dataset, which can achieve the highest recognition accuracies. In the experiments, we randomly select 20, 25, 30, 35 images per subject for training and the rest for testing. The recognition accuracies of different methods on this database are shown in Table \ref{Table_1}. Note that the mean classification accuracies and the corresponding standard deviations (acc$\pm$std) are reported, and the bold numbers suggest the highest recognition accuracies. From Table \ref{Table_1}, it is easy to find that our method can consistently achieve the highest recognition results, and outperforms the other eleven competing methods significantly, even when using a small number of training samples. Moreover, the experimental results also validate that our method has an outstanding capability on overcoming the challenges of illumination and expression variations.

\textbf{\emph{The CMU PIE Database:}} The CMU PIE face database contains more than 40,000 face images of 68 individuals in total. In our experiments, we utilize the images under five near frontal poses (C05, C07, C09, C27 and C29), and then about 170 image samples are obtained for each individual. 
We randomly select 20, 25, 30, 35 images from each subject as training samples and the remaining images are regarded as test samples. Each image is cropped and resized to be only $32 \times 32$ pixels. The detailed comparison results obtained using different methods are summarized in Table \ref{Table_2}. We can see that, with different numbers of training samples per class, our results are always better than those of all the other state-of-the-art methods, which demonstrates the effectiveness of our method.

\begin{table}[!t]
\begin{center}
\caption{Recognition accuracies (mean$\pm$std \%) of different methods with different numbers of training samples on the CMU PIE database.} \label{Table_2}
\begin{tabular}{|c|c|c|c|c|}
\hline
Alg. & 20 & 25 & 30 & 35\\
\hline \hline
LRC &90.07$\pm$0.52&92.65$\pm$0.38&94.11$\pm$0.26&94.88$\pm$0.17\\
CRC &92.52$\pm$0.33&93.84$\pm$0.39&94.31$\pm$0.16&95.52$\pm$0.16\\
SRC &92.14$\pm$0.29&93.65$\pm$0.38&94.51$\pm$0.28&95.86$\pm$0.24\\
LLC &91.90$\pm$0.25&93.27$\pm$0.56&94.66$\pm$0.41&95.26$\pm$0.49\\
SVM&90.69$\pm$0.73&92.78$\pm$0.68&93.19$\pm$0.51&94.10$\pm$0.30\\
RPCA&88.34$\pm$0.32&91.56$\pm$0.18&92.96$\pm$0.22&93.81$\pm$0.36\\
LatLRR&88.84$\pm$0.32&91.96$\pm$0.82&93.26$\pm$0.22&94.41$\pm$0.38\\
LRLR &85.83$\pm$0.56&86.90$\pm$0.45&87.82$\pm$0.49&88.23$\pm$0.42\\
LRRR &85.98$\pm$0.61&86.89$\pm$0.58&88.50$\pm$0.87&89.06$\pm$0.42\\
CBDS&91.81$\pm$0.62&93.50$\pm$0.73&94.45$\pm$0.77&94.90$\pm$0.68\\
LRSI&90.68$\pm$0.56&93.55$\pm$0.69&94.62$\pm$0.54&95.12$\pm$0.26\\
NNLRS&91.72$\pm$0.43&92.04$\pm$0.53& 93.55$\pm$0.40&94.38$\pm$0.39\\
SRRS&90.87$\pm$0.61&93.16$\pm$0.45&94.41$\pm$0.35&95.15$\pm$0.27\\
\hline
\textbf{BDLRR}&\textbf{94.67}$\pm$\textbf{0.31}&\textbf{95.79}$\pm$\textbf{0.29}&\textbf{96.46}$\pm$\textbf{0.15}&\textbf{96.81}$\pm$\textbf{0.14}\\
\hline
\end{tabular}
\end{center}
\end{table}

\begin{table}[!t]
\begin{center}
\caption{Recognition accuracies (mean$\pm$std \%) of different methods with different numbers of training samples on the AR database.} \label{Table_3}
\begin{tabular}{|c|c|c|c|c|c|c|}
\hline
Alg. & 11 & 14 & 17 & 20\\
\hline \hline
LRC &76.97$\pm$1.33&85.51$\pm$1.20&90.99$\pm$0.97& 94.22$\pm$0.76 \\
CRC &91.76$\pm$0.77&94.36$\pm$0.97&95.84$\pm$0.76& 96.63$\pm$0.87 \\
SRC &89.62$\pm$0.74&92.35$\pm$1.29&95.24$\pm$0.67& 96.19$\pm$0.75 \\
LLC &60.89$\pm$0.97&66.98$\pm$1.13&71.58$\pm$1.32&  73.53$\pm$2.15\\
SVM&86.30$\pm$1.33&92.03$\pm$0.77&95.19$\pm$0.88& 96.43$\pm$1.26 \\
RPCA&84.53$\pm$1.43&88.92$\pm$0.95&92.62$\pm$0.77&  94.90$\pm$0.78 \\
LatLRR&92.83$\pm$1.06&95.96$\pm$0.70&97.13$\pm$0.85& 97.78$\pm$0.56 \\
LRLR &88.93$\pm$0.86&93.33$\pm$0.73&94.92$\pm$0.68& 96.37$\pm$0.88 \\
LRRR &93.82$\pm$0.70&95.42$\pm$0.48&96.47$\pm$0.70& 96.88 $\pm$0.61 \\
CBDS&92.99$\pm$0.59&95.57$\pm$0.60&96.83$\pm$0.63& 97.49$\pm$0.82 \\
LRSI&86.93$\pm$1.00&90.02$\pm$0.76&93.27$\pm$0.97& 94.82$\pm$0.99\\
NNLRS&92.11$\pm$0.70&95.24$\pm$0.49&96.69$\pm$0.56&97.40$\pm$0.65\\
SRRS&87.53$\pm$1.00&93.33$\pm$1.04&96.22$\pm$1.03&97.17$\pm$0.54\\
\hline
\textbf{BDLRR}&\textbf{96.69}$\pm$\textbf{0.41}&\textbf{97.92}$\pm$ \textbf{0.30}&\textbf{98.72}$\pm$\textbf{0.42} &\textbf{99.03}$\pm$\textbf{0.38}\\
\hline
\end{tabular}
\end{center}
\end{table}

\textbf{\emph{The AR Database:}} The AR face database contains about 4,000 color face images of 126 subjects. For each subject, there are 26 images taken in two separate sessions under different conditions. In our experiments, we randomly choose a subset including 2600 images of 50 female and 50 male subjects.
Random face images of the AR face database\footnote{This dataset is publicly available from \url{http://www.umiacs.umd.edu/~zhuolin/projectlcksvd.html}.} are employed in our experiments. Following the implementation in \cite{ZJiang2013}, each image is projected onto a 540-dimensional feature vector with a randomly generated matrix with a zero-mean normal distribution. We randomly select 11, 14, 17, 20 images of each subject as training samples and treat the remaining images as test samples. The experimental results obtained using different recognition methods are shown in Table \ref{Table_3}. From the results shown in Table \ref{Table_3}, we know that our method still achieves the best recognition results, which also verifies the fact that the proposed method has particular potential for image recognition. It is notable that even when using smaller number of training samples, the performance gain of our method is still obvious in comparison with other methods.

\begin{table}[!t]
\begin{center}
\caption{Recognition accuracies (mean$\pm$std \%) of different methods with different numbers of training samples on the LFW database.} \label{Table_4}
\begin{tabular}{|c|c|c|c|c|}
\hline
Alg. & 5 & 6 & 7 & 8\\
\hline \hline
LRC &29.48$\pm$1.48&33.63$\pm$1.76&35.57$\pm$1.89&37.63$\pm$1.99\\
CRC &29.64$\pm$1.22&31.79$\pm$1.52&32.96$\pm$1.32&33.86$\pm$1.55\\
SRC &29.13$\pm$1.27&32.25$\pm$1.55&33.46$\pm$2.10&36.51$\pm$2.24\\
LLC &27.63$\pm$1.62&29.58$\pm$1.39&31.16$\pm$1.28&31.94$\pm$0.88\\
SVM&30.72$\pm$1.57&33.36$\pm$1.70&36.46$\pm$1.42&37.73$\pm$1.45\\
RPCA&31.55$\pm$1.27&34.17$\pm$1.65&36.68$\pm$1.88&37.99$\pm$1.36\\
LatLRR&30.00$\pm$1.11&33.09$\pm$1.95&35.33$\pm$1.91&37.28$\pm$1.68\\
LRLR &29.68$\pm$1.05&30.18$\pm$1.01&34.55$\pm$1.82&35.39$\pm$2.17\\
LRRR &30.98$\pm$1.28&32.93$\pm$1.70&34.86$\pm$1.03&36.59$\pm$1.87\\
CBDS&34.77$\pm$1.46&36.54$\pm$1.81&37.50$\pm$1.56&38.53$\pm$1.79\\
LRSI&31.57$\pm$2.10&34.42$\pm$1.25&37.18$\pm$0.92&39.25$\pm$1.58\\
NNSLR& 34.59$\pm$0.92 & 35.51$\pm$1.49 & 36.83$\pm$0.93 & 39.96$\pm$1.53 \\
SRRS& 31.67$\pm$1.54 & 34.29$\pm$1.74 & 38.06$\pm$1.59 & 39.43$\pm$1.65 \\
\hline
\textbf{BDLRR}&\textbf{37.83}$\pm$\textbf{1.00}&\textbf{40.94}$\pm$\textbf{1.78}&\textbf{43.11}$\pm$\textbf{1.45}&\textbf{44.51}$\pm$\textbf{1.15}\\
\hline
\end{tabular}
\end{center}
\end{table}

\textbf{\emph{The LFW Database:}} The Labeled Faces in the Wild (LFW) face database is designed for the study of unconstrained identity verification and face recognition. It contains more than 13,000 face images from 1680 subjects pictured under the unconstrained conditions. In our experiments, we employ a subset including 1251 images from 86 people, and each subject has only 10-20 images \cite{WangSJ2012} with an imbalanced number of samples. Each image was manually cropped and resized to 32 $\times$ 32 pixels. 
In our experiments, we randomly select 5, 6, 7 and 8 images of each subject as training samples and the remaining face images are treated as test samples. The experimental results of different recognition methods on this dataset are presented in Table \ref{Table_4}. We can see that the best recognition results are still achieved by our BDLRR method. Especially, the performance of our method has exceedingly advantages for this dataset in comparison with the rest of methods.


\subsection{Experiments for Character Recognition}\label{ExChar}

In this section, we evaluate the performance of our method for character recognition. More specifically, three character image datasets are employed for our experiments, including one handwriting dataset (i.e. the USPS \cite{USPS} dataset) and two scene character recognition datasets (i.e. the Char74K \cite{Char74K} and SVT \cite{SVT} datasets). It is worth noting that this work for the first time learns discriminative data representations for scene character recognition.

\subsubsection{Handwriting image recognition}

\textbf{The USPS Database}\footnote{In this study,
the publicly available set is from \url{http://cs.nyu.edu/~roweis/data.html} is used.} refers to numeric data images cropped from the scanning of handwritten digits from envelopes. It consists of 9,298 handwritten digits (`0'-`9').
All the images are resized into $16 \times 16$ pixels with 8-bit grayscale images. Each digit has about 1,100 images. In the experiments, we randomly choose 30, 60, 90 and 120 images of each digit as training samples, and regard the rest of images as test samples. The experimental results of different methods with varying numbers of training samples are shown in Table \ref{Table_5}.  The proposed method performs consistently better than all the compared methods, which further confirms that the proposed method has apparent advantages on recognizing handwriting digit images.

\begin{table}[!t]
\begin{center}
\caption{Recognition accuracies (mean$\pm$std \%) of different methods with different numbers of training samples on the USPS database.} \label{Table_5}
\begin{tabular}{|c|c|c|c|c|}
\hline
Alg. & 30 & 60 & 90 & 120\\
\hline \hline
LRC & 89.53$\pm$0.40 & 92.68$\pm$0.29 & 94.17$\pm$0.22 & 94.94$\pm$0.13 \\
CRC & 89.53$\pm$0.63 & 90.79$\pm$0.30 & 91.47$\pm$0.32 & 91.71$\pm$0.23 \\
SRC & 90.06$\pm$0.61 & 93.46$\pm$0.22 & 94.87$\pm$0.25 & 95.38$\pm$0.28 \\
LLC & 91.30$\pm$0.46 & 93.72$\pm$0.23 & 94.78$\pm$0.22 & 95.42$\pm$0.28 \\
SVM& 90.77$\pm$0.70 & 92.67$\pm$0.33 & 93.59$\pm$0.25 & 94.01$\pm$0.24 \\
RPCA& 90.07$\pm$0.29 & 93.54$\pm$0.34 & 94.72$\pm$0.12 & 95.38$\pm$0.20 \\
LatLRR& 88.75$\pm$0.70 & 90.26$\pm$0.55 & 91.08$\pm$0.34 & 91.56$\pm$0.33 \\
LRLR & 84.71$\pm$2.02 & 87.91$\pm$0.82 & 88.17$\pm$0.76 & 88.66$\pm$0.47 \\
LRRR & 86.02$\pm$2.16 & 88.22$\pm$0.84 & 88.38$\pm$0.69 & 88.77$\pm$0.45 \\
CBDS& 87.80$\pm$0.69 & 89.46$\pm$0.52 & 90.46$\pm$0.24 & 91.54$\pm$0.19 \\
LRSI& 90.62$\pm$0.41 & 93.51$\pm$0.31 & 94.54$\pm$0.17 & 95.39$\pm$0.18 \\
NNSLR& 90.54$\pm$0.57 & 93.00$\pm$0.35 & 94.03$\pm$0.22 & 94.88$\pm$0.33 \\
SRRS& 91.13$\pm$0.20 & 92.93$\pm$0.36 & 93.94$\pm$0.21 & 94.44$\pm$0.20 \\
\hline
\textbf{BDLRR}&\textbf{92.90}$\pm$\textbf{0.32}&\textbf{95.08}$\pm$\textbf{0.27}&\textbf{95.91}$\pm$\textbf{0.25}&\textbf{96.41}$\pm$\textbf{0.25}\\
\hline
\end{tabular}
\end{center}
\end{table}

\subsubsection{Scene character image recognition}: Two scene character image datasets are utilized for measuring the effectiveness of our method. As we know, natural scene character recognition is a typical yet challenging pattern recognition task due to the cluttered background, which is very difficult to separate from text. We evaluate the performance of our method in comparison with the state-of-the-art methods experimented on both datasets, including CoHOG \cite{CoHOG}, ConvCoHOG \cite{CovCoHOG}, PHOG \cite{PHOG}, MLFP \cite{MLFP}, RTPD \cite{RTPD}, GHOG \cite{GLHOG}, LHOG \cite{GLHOG}, HOG+NN \cite{SVT}, SBSTR \cite{SBSTR} and GB \cite{Char74K} (GB+SVM, GB+NN). All the images in the experiments are first resized into $32 \times 32$ pixels, and gray scale images are used in all the experiments. To make fair comparisons, we directly employ the standard partitions of training and test samples for each dataset as in \cite{ZZhang2016, CoHOG, CovCoHOG}, and the state-of-the-art algorithms evaluated on respective datasets are directly cited from their original papers. For the features\footnote{The features of both datasets are publicly available at \url{http://www.yongxu.org/databases.html}.} used in our experiments, we exploit the method in a recent paper \cite{ZZhang2016} for feature extraction. Specifically, we first use RPCA \cite{RPCA} to jointly remove noisy pixels and recover clean character images from the blurred or corrupted images, and then the well-known HOG method is applied to extract gradient features from the recovered images. The obtained HOG features are utilized for recognition.

\textbf{The Char74K Database} was collected for the study of recognizing characters in images of natural scenes. An annotated database of images including English and Kannada characters were obtained from images captured in Bangalore and India. We mainly focus on the recognition of English characters and digits (i.e. `0'-`9', `A'-`Z',`a'-`Z') with 62 classes in total. 
In our experiments, a small subset is used in our experiments, i.e. Char74K-15, which contains 15 training samples and 15 test samples per class. Table \ref{Table_6} presents the recognition results of our method and several recently proposed character recognition methods. From Table \ref{Table_6}, we can see that our method can continually achieve the highest recognition results in comparison with the state-of-the-art methods. Specifically, it is easy to see that our method outperforms the second best algorithm by a large margin of three percent.

\textbf{The Street View Text (SVT) Database} was collected from Google Street View of road-side scenes. All the images are very difficult and challenging to recognize due to the large variations in illumination, character image size, and font size and style. 
The SVT character dataset, which was annotated in \cite{SVT}, is utilized for evaluating different scene character recognition methods. About 3,796 character samples from 52 categories (no digit images) are annotated for recognition. Moreover, the SVT character dataset is more difficult to recognize than the Char74K dataset. The experimental results of using different methods on the SVT dataset are summarized in Table \ref{Table_6}. For this dataset, the proposed method significantly outperforms all the other state-of-the-art methods. We can see that the proposed method (BDLRR) achieves 79\% accuracy, which improves the accuracy by 4\% in comparison with the second best competitors such as SRC used in \cite{ZZhang2016}, CoHOG\cite{CoHOG} and PHOG\cite{PHOG}.

\begin{table}[!t]
\begin{center}
\caption{Recognition accuracies (\%) of different methods on the scene character database.} \label{Table_6}
\begin{tabular}{|c|c|c|}
\hline
\multirow{2}{*}{Alg.} & \multicolumn{2}{c|} {Testing datasets Accuracy}\\
\cline{2-3} & \textsl{Char74K-15} & \textsl{SVT} \\
\hline \hline
\textbf{BDLRR} & \textbf{70}& \textbf{79} \\
RPCA+HOG+SRC \cite{ZZhang2016} & 67& 75 \\
RPCA+HOG+Linear SVM \cite{ZZhang2016} & 63 & 73 \\
RPCA+HOG+SVM(RBF) \cite{ZZhang2016} & 63& 74 \\
ConvHOG \cite{CovCoHOG} & -& 75 \\
CoHOG \cite{CoHOG}  & -& 73 \\
PHOG (Chi-Square Kernel) \cite{PHOG} & -& 75 \\
MLFP \cite{MLFP}  & 64& - \\
RTPD \cite{RTPD} & -& 67 \\
GHOG+SVM \cite{GLHOG} & 62& - \\
LHOG+SVM \cite{GLHOG} & 58& - \\
SBSTR \cite{SBSTR} & 60& 74 \\
HOG+NN \cite{SVT} & 58& 68 \\
GB+SVM \cite{Char74K} & 53& - \\
GB+NN \cite{Char74K} & 47& - \\
\hline
\end{tabular}
\end{center}
\end{table}


\subsection{Experiments for Scene Recognition}\label{ExScene}
The performance of the proposed method for scene recognition is evaluated on the fifteen scene categories database \cite{Scene15}. It contains 4485 scene images falling into 15 categories including livingroom, bedroom, mountain, outdoor street, suburb, industrial, kitchen, opencountry, coast, forest, highway, insidecity, tallbuilding, office and store. 
The features\footnote{In this experiment, the features used are publicly available at \url{http://www.umiacs.umd.edu/~zhuolin/projectlcksvd.html}.} of fifteen scene categories provided in \cite{ZJiang2013} is employed for recognition. More specifically, the obtained features are processed as the following steps. First, the spatial pyramid feature with a four-level spatial pyramid \cite{Scene15} is computed on a SIFT-descriptor codebook with a size of 200, and then the spatial pyramid features are reduced to 3,000 by exploiting PCA to make feature dimension reduction. Following the same experimental setting of \cite{ZJiang2013}\cite{Scene15}, we randomly select 100 images per category as training data, and regard the remaining samples as test samples. For LLC, the numbers of local bases of LLC$^*$ and LLC are set to 30 and 70 respectively, which are the same parameters used in \cite{LLC2010}\cite{ZJiang2013}. Similar to above experiments, we also report the mean recognition results (mean$\pm$std) of our method over 10 times run. For fair comparison, we directly cite the results reported in LC-KSVD \cite{ZJiang2013} for performance evaluation. The experimental results are summarized in Table \ref{Table_7}. There is no doubt that our approach maintains the highest recognition accuracies and outperforms all the competing methods. Specifically, at least three percent improvements are achieved when comparing with the other methods.

\begin{table}[!t]
\begin{center}
\caption{Recognition accuracies (mean $\pm$ std \%) of different methods on the fifteen scene categories database.} \label{Table_7}
\begin{tabular}{|c|c|c|c|c|}
\hline
Alg. & Accuracy && Alg. & Accuracy\\
\hline \hline
LLC & 79.4       && SVM & 93.6\\
LLC$^*$ & 89.2   && LRSI & 92.4\\
LRC & 91.9       && CBDS& 95.7\\
CRC & 92.3       && LRRC \cite{YZhang2013}& 90.1 \\
SRC & 91.8       && SLRRR \cite{YZhang2013}& 91.3\\
LRLR & 94.4      && SRRS & 95.9 \\
LRRR & 87.2      && Lazebnik \cite{Scene15} &81.4\\
RPCA & 92.1      && Lian \cite{XLian2010}  & 86.4\\
NNLRS & 96.4      && Yang \cite{JCYang2009} & 80.3\\
LatLRR& 91.5     && Boureau \cite{YBoureau2010} & 84.3\\
LC\_KSVD1 \cite{ZJiang2013} &90.4 && Gao \cite{SGao2010} & 89.7\\
LC\_KSVD2 \cite{ZJiang2013} & 92.9&& \textbf{BDLRR}&\textbf{98.9} $\pm$ \textbf{0.19}\\
\hline
\end{tabular}
\end{center}
\end{table}
%

\subsection{Experimental Analysis}
Based on the numerical experimental results shown in Table \ref{Table_1}-\ref{Table_7}, the following observations are reached.

First, the proposed BDLRR method gains the best performances in comparison with all of the compared state-of-the-art methods for recognition tasks on eight data sets. This demonstrates that the proposed method enables to effectively learn a discriminative and robust representation from data. Moreover, we can conclude that it is beneficial to image recognition when transferring the original image features to the discriminative BDLRR based on the pivot features, i.e. training features, in a semi-supervised manner.

Second, the proposed BDLRR is significantly superior to some related methods, i.e., RPCA, LRSI, LatLRR, LRLR, LRRR, and CBDS, which demonstrates the benefit and necessity of imposing the discriminative structure on LRR and leveraging the $l_{21}$-norm to overcome noise and outliers. With the purpose of constructing the discriminative structure, the margin between block-diagonal and off-block-diagonal components is enlarged such that the incoherent data representation is boosted and the coherent data representation is enhanced simultaneously. Furthermore, it is also revealed that jointly learning the training and test representations can greatly improve the performance of recognition tasks.

\begin{figure}[!t]
\centering
\subfloat[SRC]
{\includegraphics[width=1.3in]{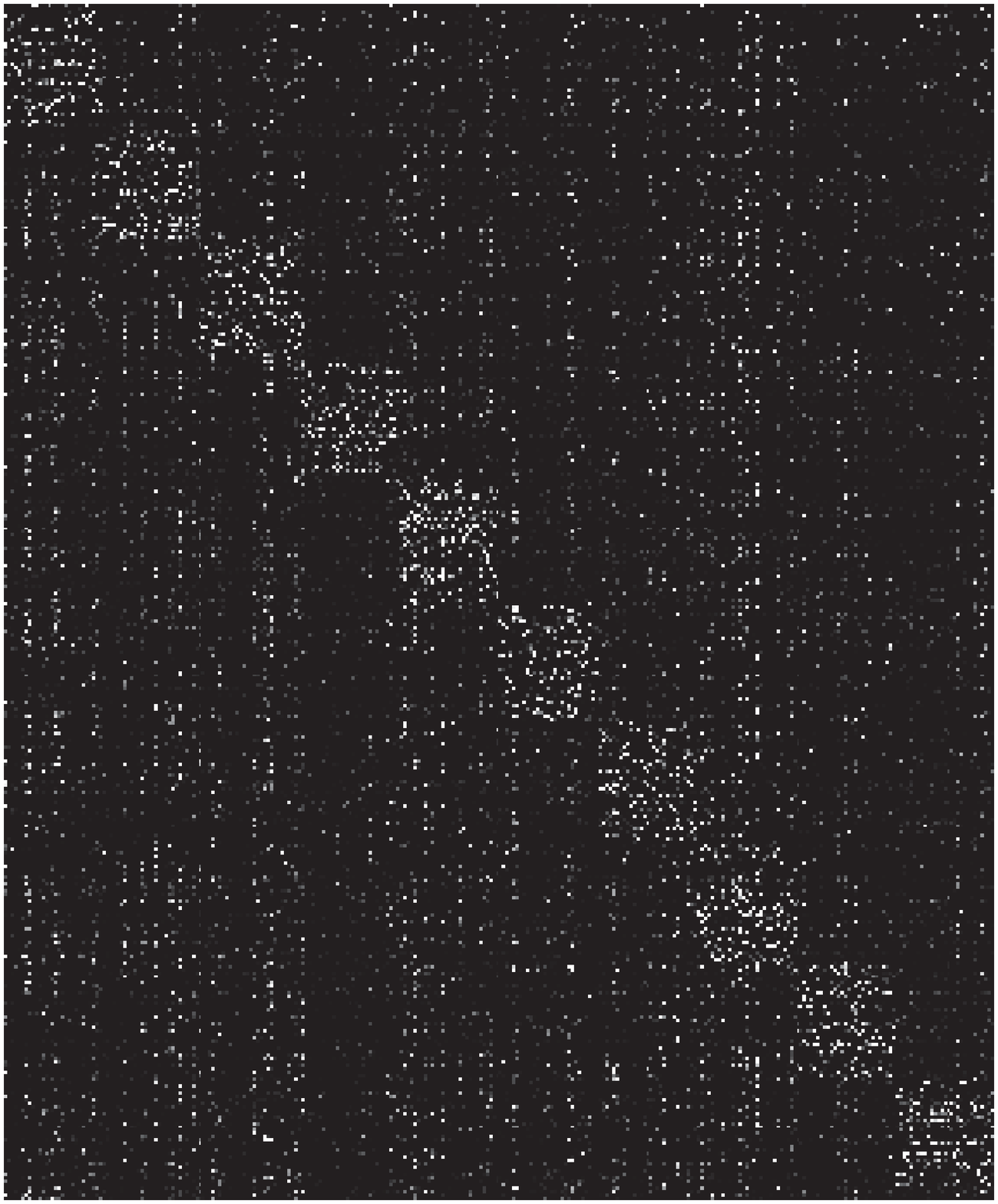}}~
\subfloat[BDLRR]
{\includegraphics[width=1.3in]{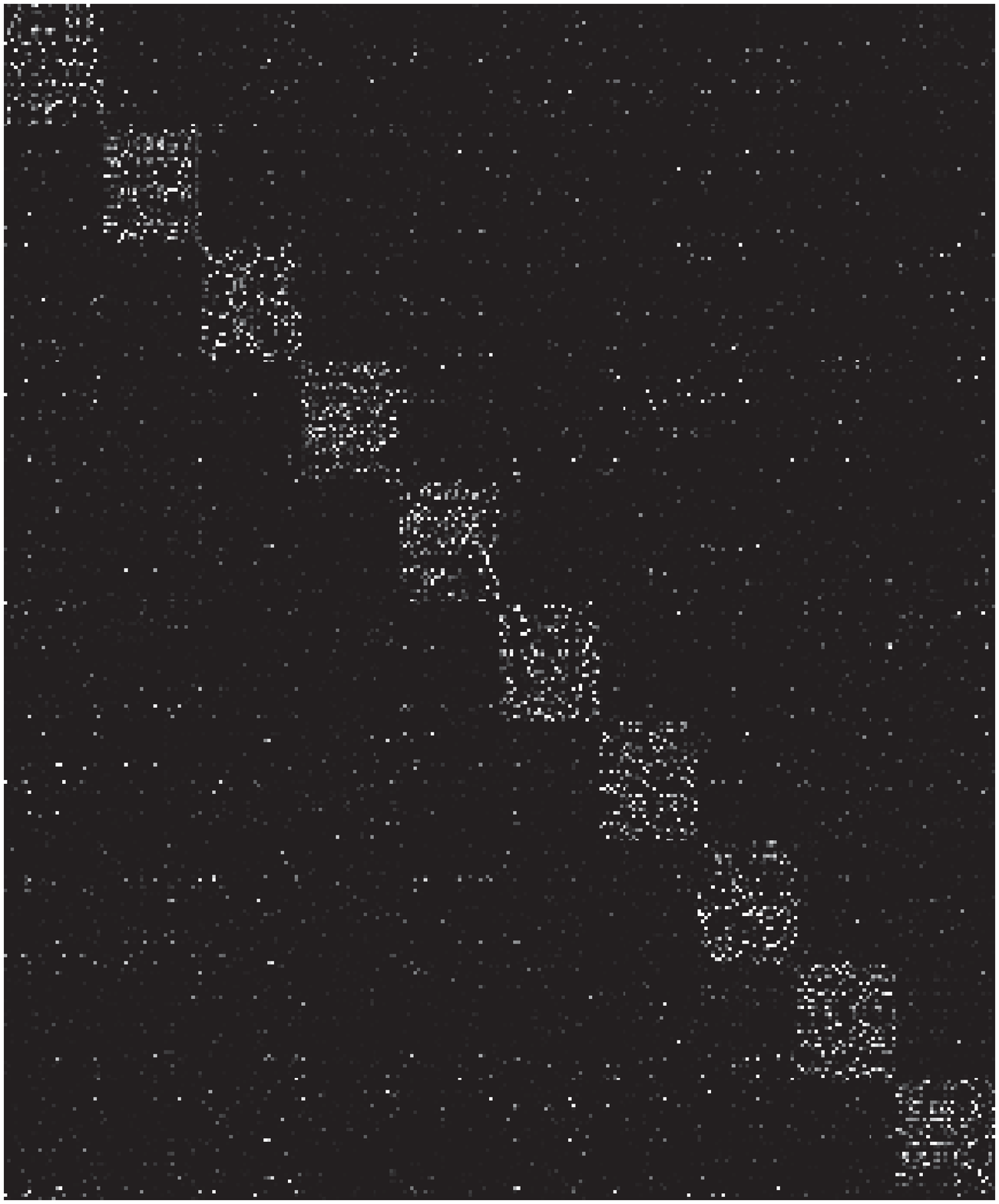}}
\caption{Data representation comparisons on the Extended YaleB dataset. (a) and (b) are data representations of the test set obtained using SRC and BDLRR, respectively. The data representation values are multiplied by 5.}\label{fig_5}
\end{figure}

Third, an interesting scenario in the experimental results is that there does not exist the absolute best algorithm among all the compared methodologies on eight datasets, because the performance relative to each other is mixed and inconsistent for different recognition applications. However, our BDLRR method outperforms all other methods on these low-resolution, limited training sample experiments. The main reason may be that our method intrinsically inherit the superiorities of sparse, low-rank, structured and elastic-net representation learning techniques. Specifically, the low-rank regularization, on the one hand, can effectively mine the underlying structure of data correlation, and the global latent structure of the data matrix is uncovered. One the other hand, the sparsity characteristic mainly focuses on finding the nearest subspace of data. However, they neglect the fact that constructing block-diagonal representation is the most straightforward fashion to explore the intrinsic structure of data and elucidate the nearest subspace of data points. For instance, we use the first 10 classes of test samples from the Extended YaleB dataset to visually present the representation results of SRC and BDLRR, which are shown in Fig. \ref{fig_5}. Images of the first ten subjects from the Extended YaleB dataset are used for experiments. We randomly select 35 images per subject as training samples and treat the rest of images as test samples. All images are rearranged by Assumption 1. From Fig. \ref{fig_5}, we can see that our method can more clearly illustrate the nearest subspace (block-diagonal structure) of test samples, leading to better recognition results.

\begin{figure*}[!t]
\centering
\subfloat[Extended YaleB ($\#Tr$ 35)]
{\includegraphics[width=1.5in]{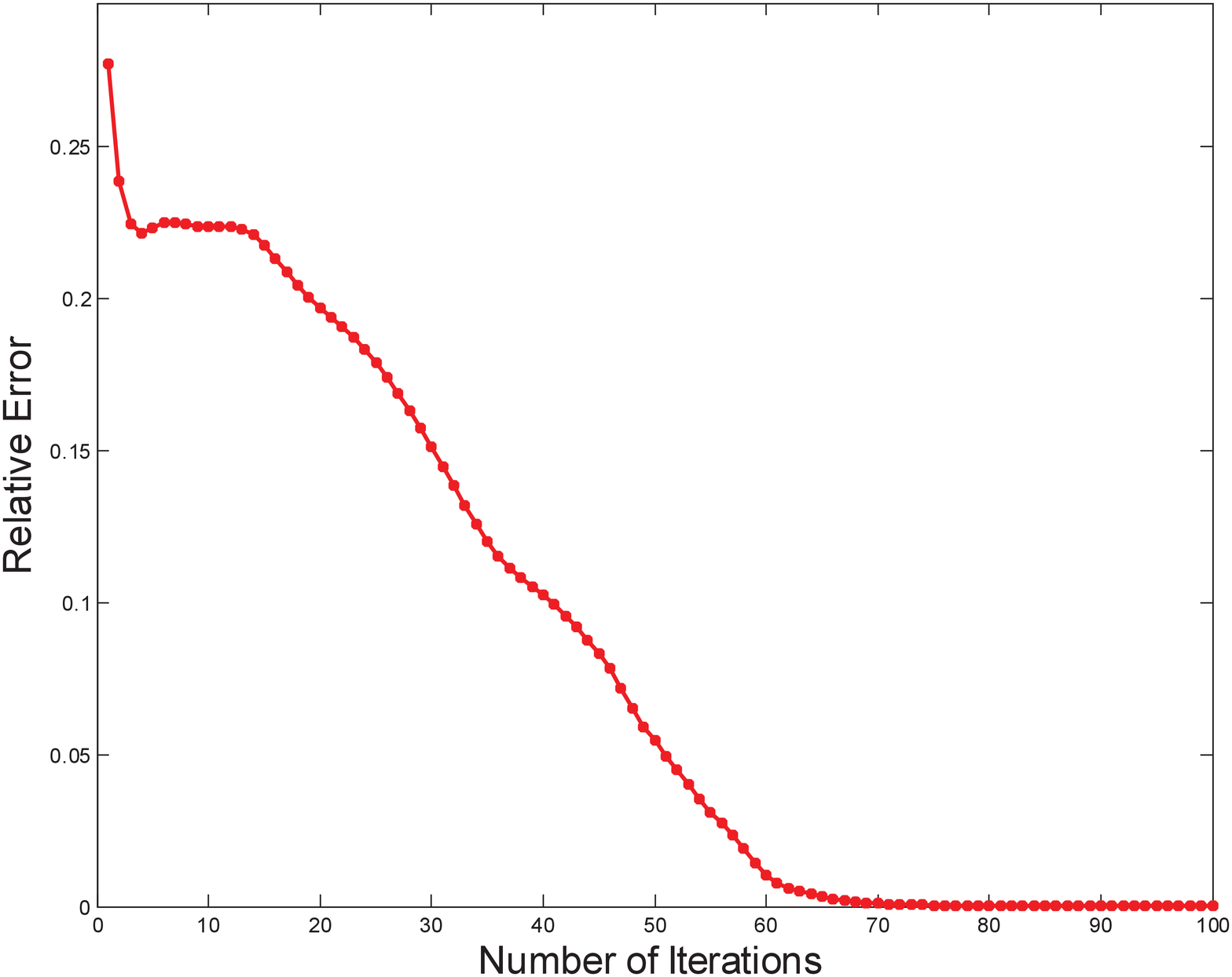}}~
\subfloat[AR ($\#Tr$ 20)]
{\includegraphics[width=1.5in]{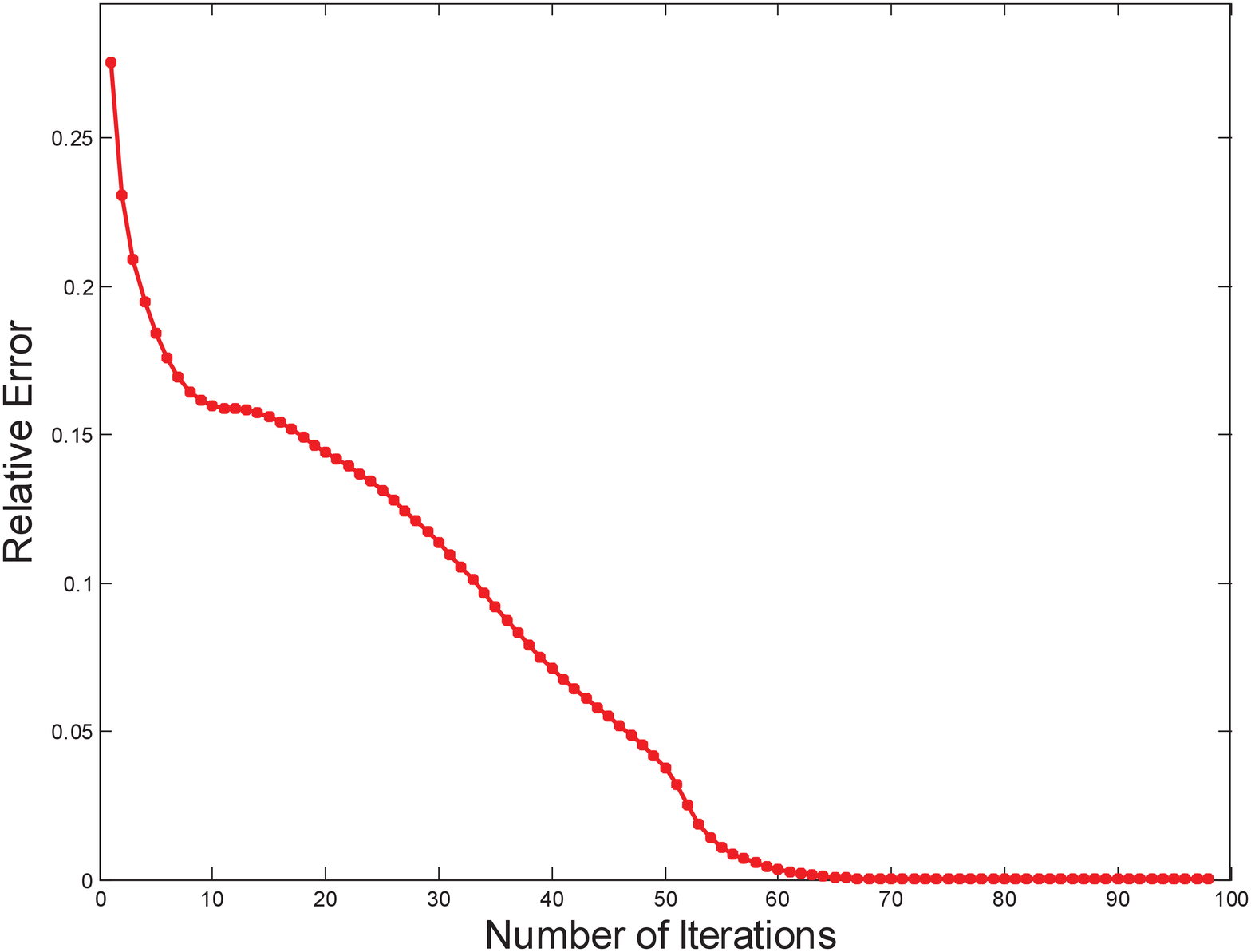}}~
\subfloat[USPS ($\#Tr$ 90)]
{\includegraphics[width=1.5in]{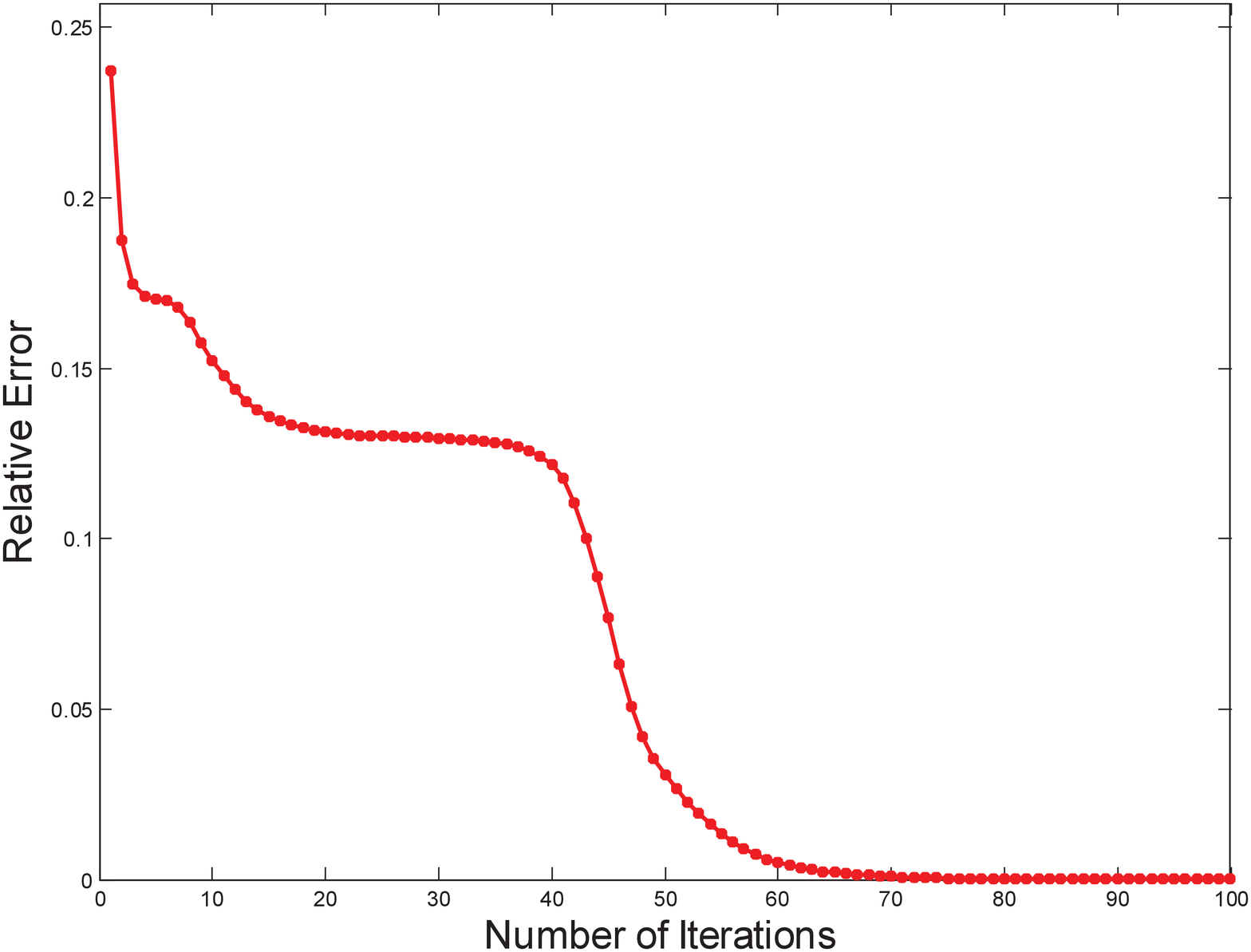}}~
\subfloat[Char74K-15 ($\#Tr$ 15)]
{\includegraphics[width=1.5in]{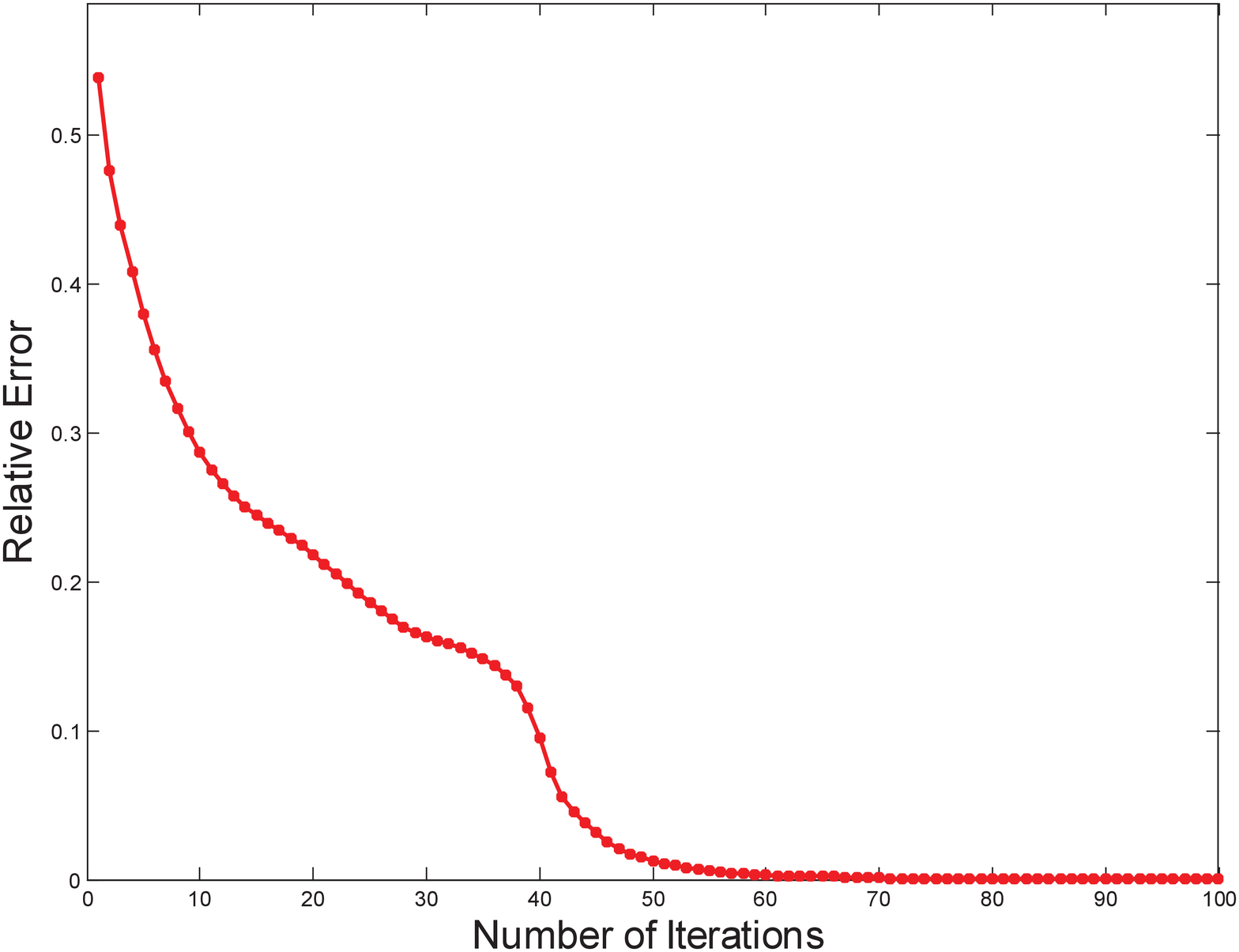}}
\caption{Convergence curves of the proposed method on different databases. (a)-(d) are the convergence curves on the Extended YaleB, AR, USPS and Char74K-15 datasets, respectively.}\label{fig_6}
\end{figure*}

\begin{figure*}[!t]
\centering
\subfloat[Extended YaleB ($\#Tr$ 35)]
{\includegraphics[width=1.5in]{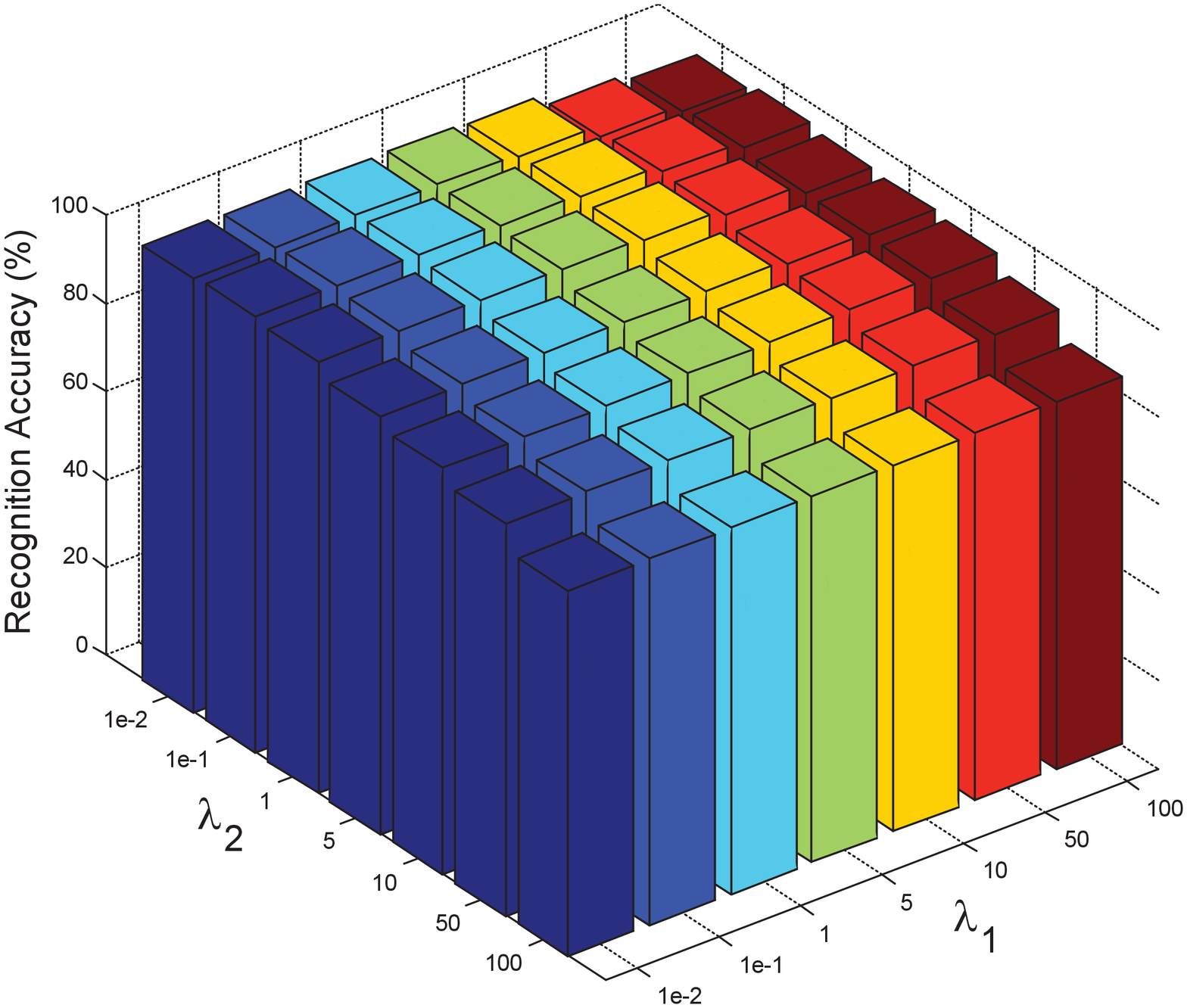}}~
\subfloat[AR ($\#Tr$ 20)]
{\includegraphics[width=1.5in]{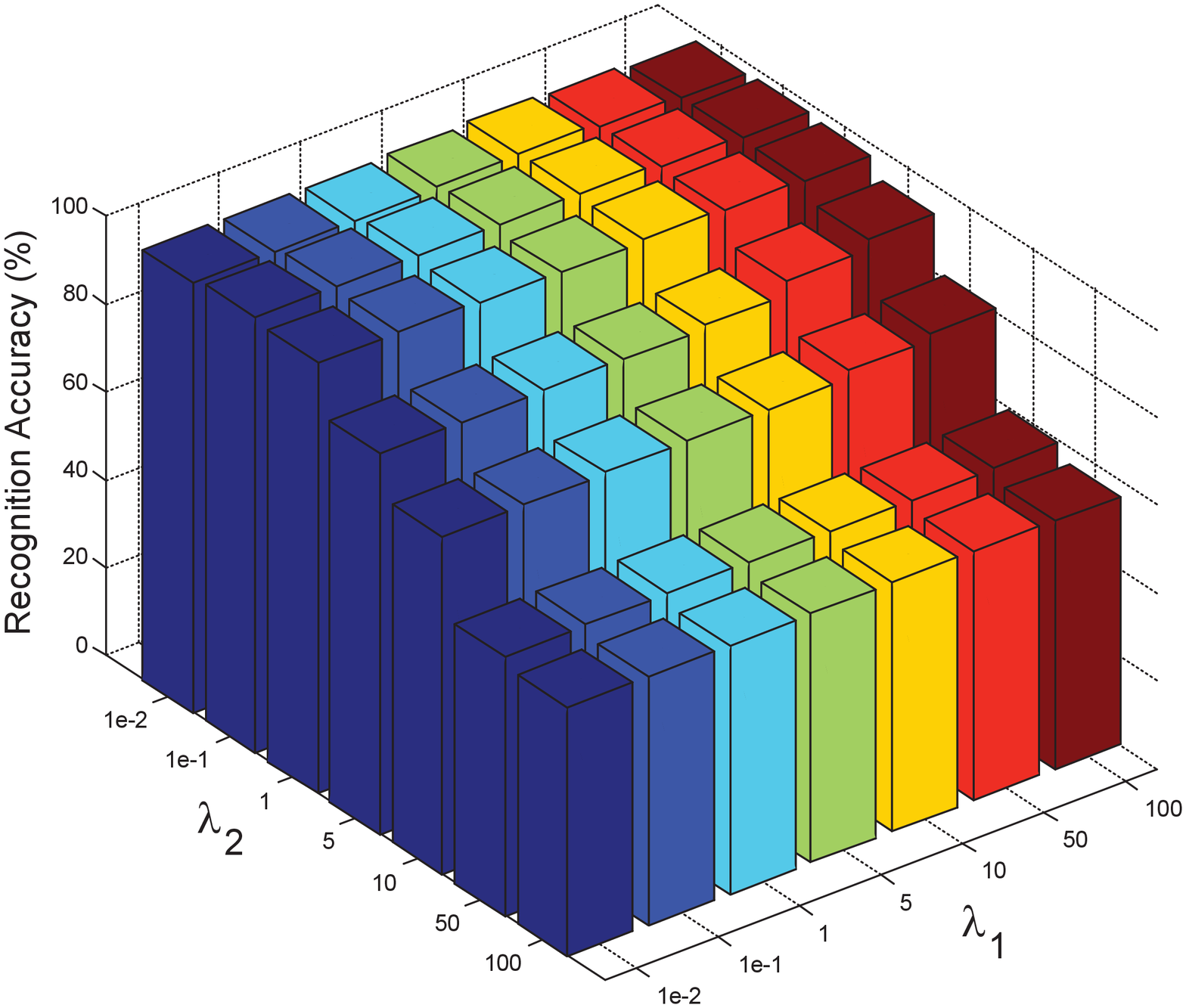}}~
\subfloat[USPS ($\#Tr$ 90)]
{\includegraphics[width=1.5in]{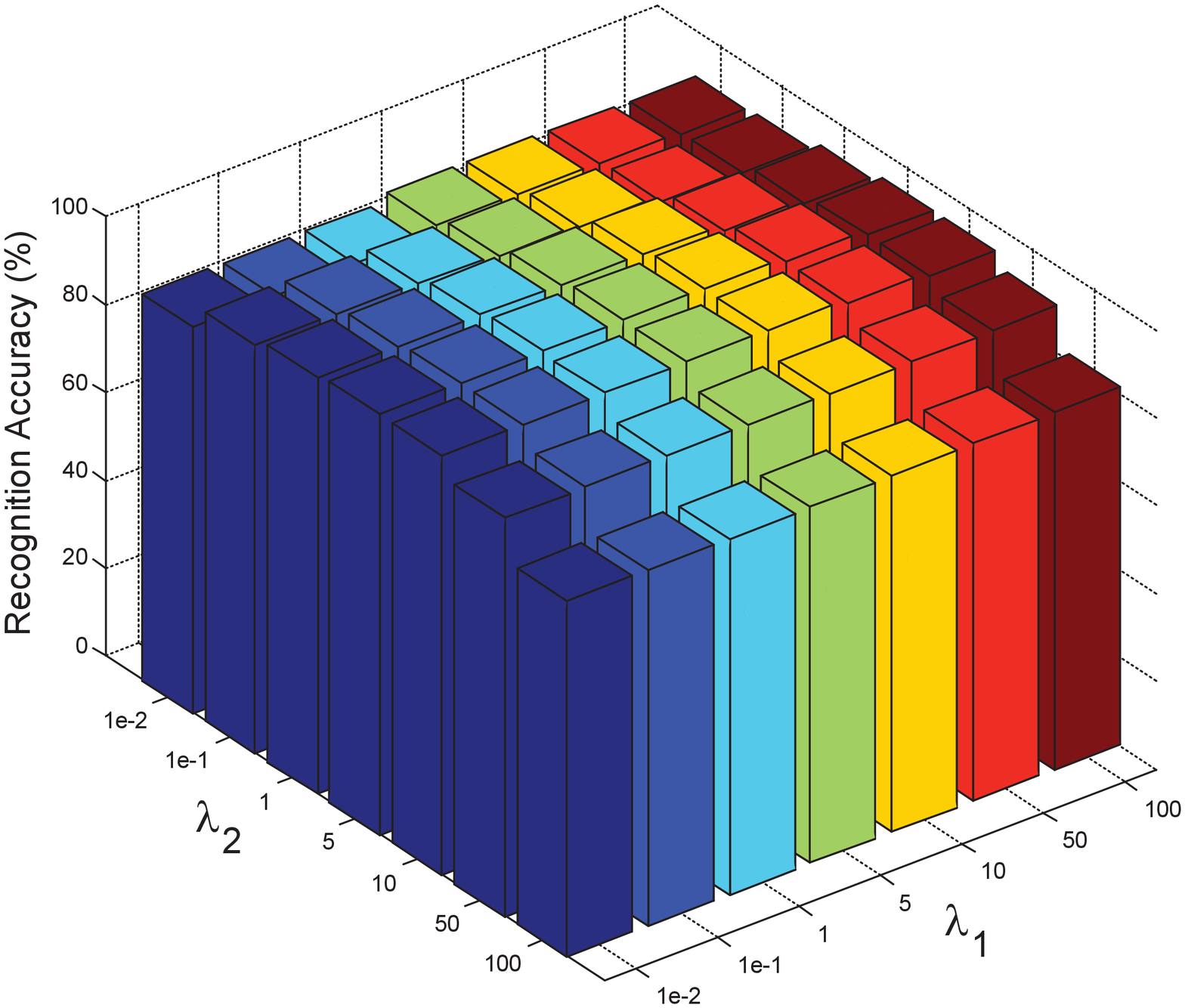}}~
\subfloat[Char74K-15 ($\#Tr$ 15)]
{\includegraphics[width=1.5in]{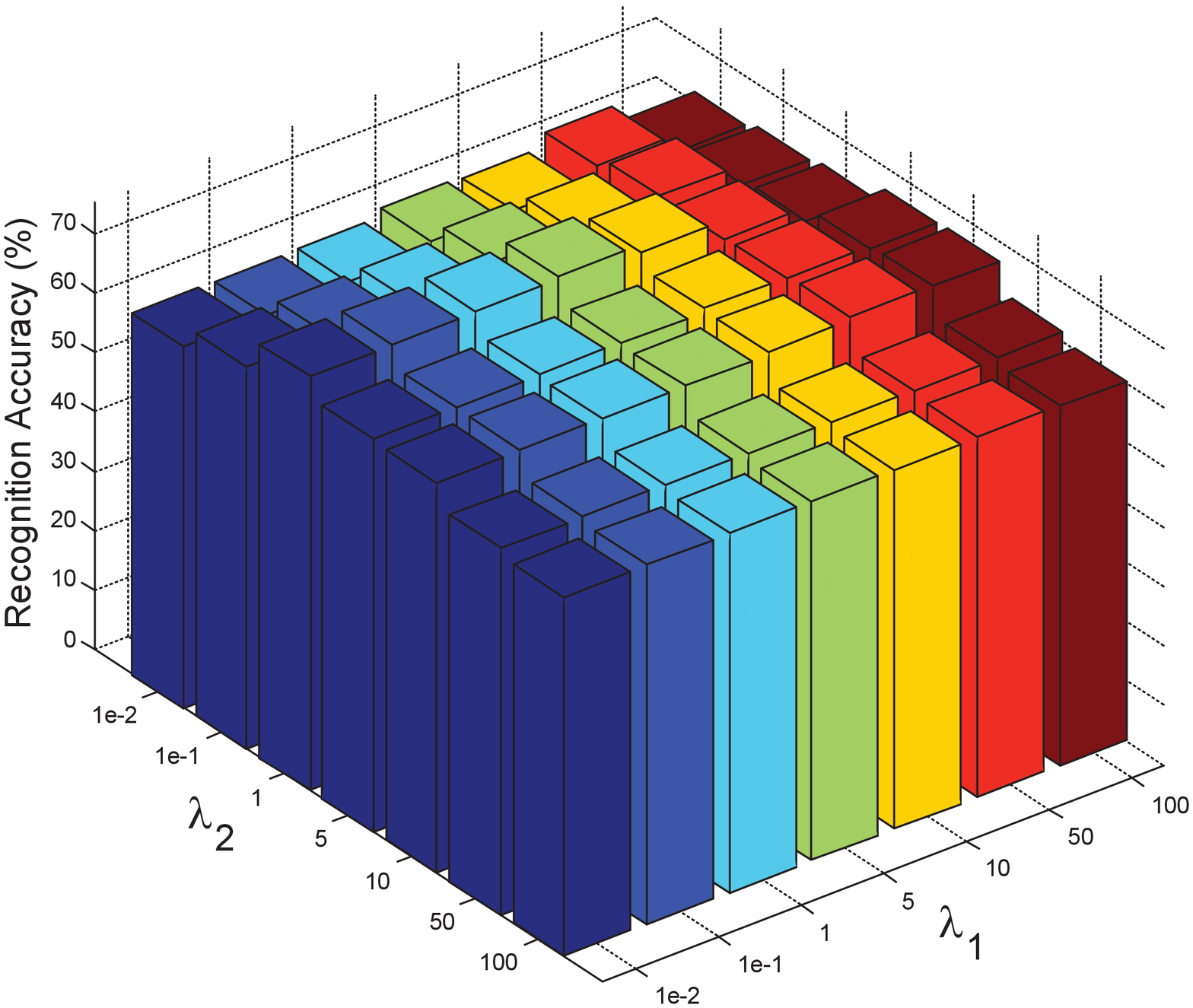}}
\caption{The performance evaluation (\%) of BDLRR versus parameters $\lambda_1$ and $\lambda_2$ on (a) Extended YaleB (b) AR (c) USPS and (d) Char74K-15 datasets.}\label{fig_7}
\end{figure*}

Fourth, our BDLRR method consistently outperforms CBDS and LatLRR on all datasets. For CBDS, it locally enforces the class-wise diagonal structure on the low-rank criterion, whereas our BDLRR method globally imposes the block-diagonal constraint on the low-rank criterion by directly minimizing the off-block-diagonal components. Moreover, our method further enhances the within block-diagonal structure to be more compact by increasing the coherent intra-class representation. For LatLRR, it extracts salient features from observations for recognition, and the unfavorable performance may result from the certain truth that too many image details are lost.

Finally, we can see that the proposed method can overcome the difficulty of noise-induced data uncertainty in face recognition, such as occlusion, disguise, severe illumination changes and expression variations. Moreover, our method works well on the challenging natural scene character recognition task, which further indicates that BDLRR is robust to the obstacles and difficulties of scene text images, such as complex background, low-resolution, occlusion, blurring, and the changes of text size or font.

\subsection{Convergence and Parameter Sensitiveness Analysis}\label{convergence}
In this section, the convergence property of BDLRR and the influence of parameter selection are empirically studied on four data sets, i.e. the extended YaleB, AR, USPS and Char74K-15 datasets.

\subsubsection{Convergence study}
The theoretical convergence proof of the proposed optimization method is analyzed in Section \ref{convProof}. It is demonstrated that BDLRR can converge to a stationary point under mild conditions. Now we experimentally validate its convergence on different datasets to demonstrate its efficient convergence. The convergence curves on four datasets are presented in Fig. \ref{fig_6}, where $\#Tr$ denotes the number of training samples per subject selected for experiments. Similar to \cite{SLi2016}, the relative error (i.e. $||X-X_{tr}Z-E||_F/||X||_F$) is employed to show its convergence. We can see that the relative error generally decreases with the increasing number of iterations. More specifically, although the relative error exists a little vibration at the first fifteen iterations on the Extended YaleB data set, the overall values of the relative error change only slightly after 60 iterations for these four datasets shown in Fig. \ref{fig_6}, which demonstrates that the proposed optimization algorithm holds the convergent nature.

\subsubsection{Parameter Sensitiveness}
In the proposed optimization problem (\ref{eq_8}), there are three parameters to be tuned. In our experiments, it is observed that the performance of BDLRR is not sensitive to $\lambda_3$ when it is in the range of [10,25], which is also an empirical setting. To test how the remaining parameters $\lambda_1$ and $\lambda_2$ influence the performance of BDLRR, we perform extensive experiments to validate their robustness. Similar to convergence validations, we still use the Extended YaleB, AR, USPS and Char74K-15 datasets for evaluation. Fig. \ref{fig_7} presents the performance variations with respect to parameters $\lambda_1$ and $\lambda_2$. We can see that the performance of our BDLRR method is generally insensitive to varying values of $\lambda_1$ and $\lambda_2$. More specifically, the performance is promising when parameter $\lambda_1$ is not too large or small, which indicates the necessity of boosting the extra-class data incoherent representation. Moreover, for parameter $\lambda_2$, it is easy to see that it should be small, and the best results are usually achieved when the value is smaller than 1, yet bigger than 0.01. The possible reason of a smaller $\lambda_2$ may be that the Euclidean distance metric used in our experiments is too simple to perfectly measure the similarity of samples. However, we have achieved very impressive experimental results, even with a simple distance metric. In a word, our BDLRR method is robust to parameter changes in most cases.

\subsection{Limitation}
From the objective function of our BDLRR method, i.e. Eqn. (\ref{eq_8}), we can see that the proposed model is a semi-supervised representation learning model and concurrently learns both block-diagonal representations of training and test samples, which indicates that the test samples and the label of training samples are both given in the learning process. However, in some cases we cannot get access of test data at the training stage, which may limit the generalization of our model. To this end, we extend our BDLRR method to address the out-of-sample problem in section \ref{OE} to circumvent this problem. In this way, our results are somewhat subject to the learning capability of algorithms in handling the out-of-sample cases. Fortunately, these methods have been examined to effectively formulate favorable representations of new instances. Moreover, the learned data representations of training samples are reasonably block-diagonal in the training stage, which in turn guarantees the satisfactory recognition results.

\section{Conclusion} \label{conc}
In this paper, we have proposed a novel discriminative block-diagonal representation learning model, i.e. BDLRR, for robust image recognition. BDLRR focuses on learning a discriminative data representation by imposing an effective structure in a low-rank representation framework, where the extra-class incoherent representation and intra-class coherent representation are simultaneously enhanced. The proposed method incorporates the learned BDLRR into the semi-supervised model to collaboratively optimize the training data representation and test data representation, and then an efficient linear classifier is obtained to perform final robust image recognition. Moreover, an effective optimization algorithm is developed to solve the resulting optimization problem. Last but not least, the proposed method was evaluated on eight publicly available benchmark datasets for three different recognition tasks. Extensive experimental results have demonstrated that the proposed BDLRR method is superior to state-of-the-art methods.

\section{Acknowledgement} \label{Acknowledge}
We would like to thank Prof. Chenglin Liu and Dr. Xuyao Zhang for many inspiring discussions and constructive suggestions. Moreover, we also thank the editor, an associate editor, and referees for helpful comments and suggestions which greatly improved this paper.

\ifCLASSOPTIONcaptionsoff
  \newpage
\fi

\end{document}